%% file: main2.tex
\documentclass[a4paper,x11names,10pt,twoside]{article}
\usepackage{manuscript}

\bibliography{ref}

\title{What Does `Human-Centred AI' Mean?}
\author[1,2]{Olivia Guest}

\affil[1]{\protect\begin{varwidth}[t]{\linewidth}\protect\centering Department of Cognitive Science and Artificial Intelligence,\par Radboud University, The Netherlands\protect\end{varwidth}} 

\affil[2]{\protect\begin{varwidth}[t]{\linewidth}\protect\centering Donders Institute for Brain, Cognition, and Behaviour,\par Radboud University, The Netherlands\protect\end{varwidth}} 

\makeatletter
\edef\mytitle{\@title}
\makeatother

\fancypagestyle{plain}{%
  \renewcommand{\headrulewidth}{0pt}%
  \fancyhf{}%
  \fancyfoot[R]{\small\thepage}%
}
\begin{document}
    % Set the page style to "fancy"...
    \pagestyle{fancy}
    \renewcommand{\headrulewidth}{0pt}
    %... then configure it.
    \fancyhead{} % clear all header fields
    \fancyhead[LO]{\small\sc\mytitle}
    \fancyhead[LE]{\small\sc O. Guest}
    \fancyfoot{} % clear all footer fields
    \fancyfoot[R]{\small\thepage}%
    
    \renewcommand{\abstractname}{}
    \renewcommand{\absnamepos}{empty} % originally center
    \renewcommand{\abstracttextfont}{\small}
    \maketitle\thispagestyle{plain}
    
    \begin{abstract} % ditto abstract
    % \vspace{-2em}
    \noindent {\normalfont\bfseries Abstract: }\input{abstract}
    
    \vspace{1em}
    \keywords{artificial intelligence; cognitive science; sociotechnical relationship; cognitive labour; artificial neural network; technology; cognition; human-centred AI} 
    \end{abstract}

\vspace{1em}

\input{body}

\printbibliography
\end{document}

%% file: abstract.tex
% While it seems sensible that human-centred artificial intelligence (AI) means centring ``human behaviour and experience'', it cannot be any other way.
% % and results in contradiction.
% AI, I argue, is necessarily seen as a relationship between technology and humans where it appears that artifacts can perform, to a greater or lesser extent, human cognitive labour.
% This is evinced using examples that juxtapose technology with cognition, inter alia: abacus versus mental arithmetic, alarm clock versus knocker-upper, camera versus vision, and sweatshop versus tailor.
% Using novel definitions and analyses, sociotechnical relationships can be analysed into varying types of displacement (harmful), enhancement (beneficial), and/or replacement (neutral) of human cognitive labour.
% Ultimately, all AI implicates human cognition; no matter what. Obfuscation of cognition in the AI context --- from clocks to artificial neural networks --- results in distortion, in slowing critical engagement, in perverting cognitive science, and indeed in limiting our ability to truly centre humans and humanity in the engineering of such systems.
% To even begin to de-fetishise AI, we must look the human-in-the-loop in the eyes.
While it seems sensible that human-centred artificial intelligence (AI) means centring ``human behaviour and experience,'' it cannot be any other way.
% and results in contradiction.
AI, I argue, is usefully seen as a relationship between technology and humans where it appears that artifacts can perform, to a greater or lesser extent, human cognitive labour.
This is evinced using examples that juxtapose technology with cognition, inter alia: abacus versus mental arithmetic; alarm clock versus knocker-upper; camera versus vision; and sweatshop versus tailor.
Using novel definitions and analyses, sociotechnical relationships can be analysed into varying types of: displacement (harmful), enhancement (beneficial), and/or replacement (neutral) of human cognitive labour. Ultimately, all AI implicates human cognition; no matter what.
Obfuscation of cognition in the AI context --- from clocks to artificial neural networks --- results in distortion, in slowing critical engagement, perverting cognitive science, and indeed in limiting our ability to truly centre humans and humanity in the engineering of AI systems.
To even begin to de-fetishise AI, we must look the human-in-the-loop in the eyes.

%% file: body.tex
\section{Introduction}\label{sec:intro}
\begin{quote}
    [M]odels have to be used before they will give up their secrets. In this sense, they have the quality of a technology --- the power of the model only becomes apparent in the context of its use. \citep[][p. 12]{Morrison_Morgan_1999}
\end{quote}
We are experiencing a point in history that considers itself separate, unique, a zenith of human civilisation, presented as disconnected from the past through a series of technological sea changes.
In extreme cases, we are detached from even the last few months: the newest version of a piece of artificial intelligence (AI) software is the definitive one, everything else is irrelevant.
And so findings in AI are held to expire.
Our AI students describe their own knowledge and skills --- even their whole degree itself --- as having upcoming sell-by dates.
Research in AI is thus often framed as unmoored from historical, ethical, social, and legal precedents.

\begin{quote}
The speed of the ``ultra rapid computing machine,''
as \citet{wiener1948,Wiener1950} describes computers, becomes a metaphor for the
speed of change and of technological progress. 
This notion of condensed time operates as a further discursive regularity in two ways. First, it forms an overall temporal backdrop against which various cybernetic dramas are played out. Second, condensed time becomes a measure of the performance of humans and machines. 

\citet{hughes1985world} offers a telling example of the first manifestation of condensed time:
\begin{quote}
    In the scant two hundred years since the early Industrial Revolution, Western man has totally rescaled and changed the face and fabric of his environment. Indeed, the changes have proceeded at such an accelerated pace that we might use the word ``old'' or ``outmoded'' to refer to last month’s computer model. (p. 205) 
\end{quote}
This sense of condensing time---of speed, of rapid change---imbues virtually
all the popular literature examined. It becomes a contextual operator, stated but not questioned. \citep[][p. 193]{Hamilton_1998}
\end{quote}
What Sheryl N. Hamilton describes above is more than seven decades old. But if this has been going for decades, two-thirds of a century or longer still, does this not undermine its very premise?
How can we both be at a unique point in history, and trapped in an endless cycle that began with the paralleling of humans and clockwork?

The process of uncondensing time indubitably requires slowing down \citep{stengers2018another} and radically centring the human and decentring the ``ultra rapid computing machine.'' \citep{Hamilton_1998}
To do this, some propose to (conceptually) reengineer \citep[][]{nado2021conceptual, guest2024makes} or reimagine these machines and our relationships to them inter alia through a subfield known as human-centred AI.
For a pertinent example, Andy J. \citet{wills2025}, in the call for this special issue, describes human-centred artificial intelligence (HCAI) as placing ``human behaviour and experience at the heart of [AI] research.''
An example of this, he says, could be cases wherein it is ``claimed that artificial neural networks (ANNs) now perform at human levels in a variety of tasks''.
He goes on to ask:
\begin{quote}
    Can ANNs effectively and safely be used to support the work of highly trained professionals—for example, radiologists, therapists, legal advisors, or researchers? Can we effectively adapt the skills and techniques of behavioural research, previously applied to humans and other animals, to better understand the ‘psychology’ of complex black-box ANNs?
\end{quote}
Relatedly, Yvonne \citet{rogers2022commentary} states the goal of HCAI as ``designing AI systems that enhance human capacities and improve their experiences rather than replacing them through automation[, along with the aspiration of] reimagining human-machine interaction in all its guises[, while negotiating] the creeping creepiness of AI'' (p. 255).
Industry players, such as IBM, make a different but not entirely incompatible conceptualisation:
\begin{quote}
    Human-Centered AI (HCAI) is an emerging discipline intent on creating AI systems that amplify and augment rather than displace human abilities. HCAI seeks to preserve human control in a way that ensures artificial intelligence meets our needs while also operating transparently, delivering equitable outcomes, and respecting privacy.
\citep[][n.p.]{Geyer_Weisz_Pinhanez_Daly_2022}
\end{quote}
Regulatory bodies, like the European Union, also have a perspective:
\begin{quote}
    The human-centric approach to AI strives to ensure that human values are central to the way in which AI systems are developed, deployed, used and monitored, by ensuring respect for fundamental rights, [...] which are united by reference to a common foundation rooted in respect for human dignity, in which the human being enjoys a unique and inalienable moral status.
    % This also entails consideration of the natural environment and of other living beings that are part of the human ecosystem, as well as a sustainable approach enabling the flourishing of future generations to come.
\citep[][p. 3]{tambiama2019eu}
\end{quote}
Overall we see that HCAI --- as a specialisation for students \citep[e.g.][]{tue, delft}, as a subfield for scholars \citep[e.g.][]{wills2025, holzinger, rogers2022commentary, uu, HCAIMConsortium, HomeStan55} or technology industry workers \citep[e.g.][]{Geyer_Weisz_Pinhanez_Daly_2022} alike, and as a concern and mission for regulators \citep[e.g.][]{pirozzoli2024human, tambiama2019eu}  --- interweaves four related themes:
\begin{enumerate}
    \item supporting or enhancing human skills, both without displacement of said skills and without violation of fundamental human rights;
    \item imbuing systems with various so-called human-aligned values, including those of explainability and transparency;
    \item focussing on human behaviour as a benchmark, i.e. the idea of human-like or -level performance;
    \item implicating behavioural, or otherwise psychological, methods in the study of these systems as if on equal epistemic footing with humans.
\end{enumerate}

In this paper, I turn this frame on its head \citep[cf.][]{bannon2011reimagining, rogers2022commentary, ryan2024we, schmager2025understanding, dyerwitherford2019, walton2021rethinking, andersen2023introduction, shneiderman2022human, bishop2021artificial, steinhoff_2021}.
Using a unique (re)definition of AI that releases us from a correlationist\footnote{The idea that it is correlations with a given target, like human data derived benchmarks (\citealt{Schrimpf2020integrative, pasquinelli2017machines}; cf. \citealt[][]{van2024reclaiming}), that validate the safety \citep[][]{el2022impossible},
appropriateness, or other desirable properties of a model or system.
In general, correlationism may be unproblematic, but in a setting where computation is involved correlations cannot function as a useful guide, serving more as red herring than anything else \citep[][]{pittphilsci24834, pittphilsci25289, guest_martin_2024, guest2023logical}.} grip, we then examine three triplets as case studies of techno-social relationships between cognition and artifacts.
% , so elucidate properties of 
To presage the coming analyses, herein AI is any techno-social relationship that outsources to machines or algorithms some part, however small, of human cognitive labour \citep[cf.][]{van2024reclaiming}.
I demonstrate that AI is human-centric, \textit{not} because it behaves like or is designed to be like humans, but because it requires a ghost in the machine, often literally an obfuscated human-in-the-loop to properly function \citep[also see][]{pittphilsci24834} because AI \textit{is} humans albeit in fetishised \citep{morris2017after, braune2020fetish, mota2024fetichismo}, obfuscated forms \citep[e.g.][]{erscoi_kleinherenbrink_guest_2023}.
That is, AI ``is in reality produced by relations among people [even though it] appears before us in a fantastic form as relations among things'' \citep[][p. 250]{pfaffenberger1988fetishised}.
AI's ``technological veil'' hides human cognitive labour \citep{mota2024fetichismo}.
I bring this anthropological, sociotechnical and broadly computational cognitive scientific angle to understanding AI, that is unlike ``the skills and techniques of behavioural research, previously applied to humans and other animals[, which claim to help us] better understand the `psychology' of complex black-box ANNs'' (\citealp[][n.p.]{wills2025}; cf. \citealp{raley2023critical}).
In fact behavioural probing of such systems using such experimental techniques assumes their psychological standing to be equal to, or comparable to, biological organisms: it begs the question \citep[see for relevant analyses:][]{guest_martin_2024, guest2023logical, forbes2025improve, raji2022fallacy}. 
\begin{table}[bht]
    \centering
    \caption{The two steps required for the proposed redefinition of AI.
    At the top is step \textit{1}, where we decide whether a relationship exists between a technology and human cognition.
    This relationship, represented by the blue-green column between \textit{Machine} and \textit{Human} on row \textit{1a}, is AI.
    In \textit{1b} are terminological examples, both non-diagnostic on their own and incomplete as a list, that can aid in the diagnosis of a sociotechnical relationship as one of AI. 
    The three columns below in step \textit{2} represent three, not mutually exclusive, types of sociotechnical relationship between humans and artifacts.
    % that I propose is a useful way to think about AI.
    % So to determine if we can analyse a technology or tool this way, we must first ask: Is the artifact in a sociotechnical relationship with human cognitive labour? If yes,
    At this step, we sketch out if AI replaces, enhances, or displaces cognition (row \textit{2a}) --- with relevant properties and their typical values, non-exhaustively specified, listed on rows \textit{2b--h}.}
    \input{table1}
    \label{tbl:1}
\end{table}

The perspective I bring here answers questions upstream to analysing the behavioural outputs of such systems, instead focussing on \textit{in principle} analyses, freeing us from a correlationist account which delivers flawed reasoning and uninterpretable results \citep{guest_martin_2024, guest2023logical, pittphilsci25289}.
The contradiction between artifice, artificiality, machines, and the machinic and intelligence, cognition, and humanity is problematised and dissected by the method herein.
These two can appear both as opposites and as identical; at odds as analytical constructs and inexorably intertwined.
For example, ``[f]or a long time, the human was something else altogether; it is not so long ago that it became a machine---a calculating one no less.'' (\citealp{mauss1923}, pp. 176--177; translated by \citealp{lepage2024}, p. 20)
Some go further along this route, remarking that ``the history of the sciences is now reaching a point, in all its branches, where every scientific theory can be taken as a machine'' \citep[][p. 112]{alma9924120787602466}; and arguing ``that scientific models have certain features which enable us to treat them as a technology.'' \citep[][p. 35]{Morrison_Morgan_1999}
And others --- coinciding with the regulators' worries --- notice that ``[m]achinery does not just act as a superior competitor to the worker, always on the point of making [them] superfluous. It is a power inimical to [them], and capital proclaims this fact loudly and deliberately, as well as making use of it.'' \citep[][n.p.]{FullMarx27:online}
Which is it?
Are humans machines, or machinic in some important way, so they can be recreated in machines?
Are machines somehow human-like, created by us in our image?
Let me tell you.

\begin{table}[tbh]
% \begin{samepage}
\caption{The first column represents the AI relationship (defined in \autoref{tbl:1}) between using an abacus  and non-abacus assisted mental arithmetic.
The second column between using an electronic or mechanical calculator and an unassisted adult who knows basic mathematical operations already.
The third column between using a digital computer and a human computer.
% , which was a profession from the Middle Ages to the middle of the previous century --- the comparison is during the period where human computers were pushed out of the industry \citep{hicks2017programmed, grier2013computers}.
The photo in the third column depicts NASA human computers Dorothy Vaughan, Lessie Hunter, and Vivian Adair \citep{DorothyV75:online, shetterly2016hidden}. 
% Each row for each AI relationship is explained in the main text.
}
\label{tbl:2}
\input{table2}
% \end{samepage}
\end{table}
\section{Radically Redefining AI}
\begin{quote}
% This experience has demonstrated that 
[I]t is impossible to create an absolutely reliable automatic system, and sooner or later people face the necessity to act after equipment fails. [...]
% The cosmonaut must be constantly prepared to take up the functions of the failed system. But if his functions are limited to monitoring and observation only, then he is effectively excluded from the control process. To be able to join in the control process, he must have strong manual control skills based not only on his experience with ground tests, but also on his performance of control functions in real flight conditions.
If the cosmonaut loses such skills because of [their] passive role [due to being typically limited to monitoring and observation only], the probability of [their] choosing and carrying out the right procedure in an emergency would be small. This contradiction is inherent in automatic control systems.
% More here: https://web.mit.edu/slava/space/essays/essay-ponomareva.htm and: https://web.mit.edu/slava/space/interview/interview-ponomareva.htm
\citep[][n.p.]{ponomareva1998human}
\end{quote}
As practitioners of science, we are duty bound to  
% Perhaps this is unsurprising when
think about what `artificial' and `intelligence' mean.
What does it mean to propose an artificial version of a human capacity?
% This parsimonious account will be analysed in depth below, but we need to address some preliminaries before that.
When we talk about redefining something, we must remember that ``definitions have no inherent truth.
They are agreed conventions tested by their internal coherence as well as by their relationship to common sense, common practice and history.'' (\citealt[][p. 790]{yuval2024antisemitism}; also see \citealt{elgin2017true, nado2021conceptual})
Some of the existing definitions of AI when probed reveal weaknesses in the aforementioned dimensions --- especially undesirable when speaking of formal systems, which all AI systems necessarily are.
This is to say, ``weakness from the viewpoint of formal symmetry of doctrine [is] strength in the service of rising capitalism.'' \citep[p. 25]{novack1968empiricism}
So these formal and common-sense problems with pre-existing AI conceptions form a core of strength when incentives are, in opposition to academia, ones of profit-making for the technology industry (see \cite{van2024reclaiming}, for an exposition of possible meanings).
AI has had so many related but nonetheless different meanings over the decades that perhaps it is a fool's errand to even try and pin it down \citep{guest_martin_2024, boden2006, lighthill1973artificial, haigh2023there, dreyfus1965alchemy, mccorduck2004machines, smith_smith_2024}.
And similar issues with terminology appear with related terms like `brain inspired' and `neurally plausible' to name but a few terms which function as weasel words \citep{guest_martin_2024} --- as well as when `computational' is used metaphorically or naively by cognitive neuroscientists \citep{pittphilsci24834, pittphilsci25289}.

All AI implicates human cognition, as a user, as a human-in-the-loop, and as an inspiration.
What I propose is that the important aspect is understanding how it does this; understanding the sociotechnical \textit{relationship}, as opposed to understanding (e.g. probing, benchmarking) the technology in the abstract. 
% \citep[][]{paper on myserianism}.
The shifting of the scientific emphasis to the relationship --- what does the model do as a function of interacting with humans? --- from what the model is in and of itself is by no means completely alien to cognitive scientists \citep[][]{BAINBRIDGE1983775}.
For example, cognitive computational modellers are familiar, consciously or otherwise, with what in the philosophy of science is known as the pragmatic view on scientific theories and models, i.e. models are characterisable by their properties derived from their use  \citep[][]{morgan1999models, sep-structure-scientific-theories, guest2021computational}.

Without further ado and notwithstanding terminological disarray, \autoref{tbl:1} embodies a single tripartite definition that acts as an analytical tool for discerning the AI's properties, e.g. labour obfuscation, which is the extent of the hidden human-in-the-loop \citep{pittphilsci24834, crawford2021atlas}.
% "Hobbes went much further. He
% tried to squeeze all types of movement, from the opera-
% tions of nature to social and political actions and the process of the human mind, into the framework of purely mechanical categories." p.28--29 novack
% Returning to the view on Ai as inherently sociotechnical
% AI is a relationship between technology and humans where it appears that artifacts can perform, to a greater or lesser extent, human cognitive labour.
The radical redefinition I propose comprises two parts.
% , i.e. when we analyse if something is AI under my conception our thought process can be decomposed into two steps.
First, full-blown deflation --- any sociotechnical relationship could be AI if it links an artefact to human cognition such that the artefact can be seen as performing some aspect of a cognitive capacity \parencite[cf.][]{egan2025deflating, pittphilsci24834}.
For this step, all we need to ask ourselves is: does the technology exist in a relationship with human cognition? If we affirm this, we can move to the next step.
Second, we reinflate AI into three types of relationship, that of enhancement, of replacement, and of displacement (see \autoref{tbl:1}) of human cognitive labour. 
% We will return to step two in the next section.

As mentioned, AI is any relationship between technology, tools, models, machines and humans where it appears as if cognitive labour is offloaded onto such artifacts.
Anthropological notions of tools versus technologies can help to unpack what is going on with respect to such artifacts:
% --- although I am not taking a stand as to whether or not AI as a relationship excludes tools:
\begin{quote}
Tools are creations on a localised, small-scale, the products of either individuals or small groups on specific occasions. As such, they do not give rise to systems of control and coercion. Technology, on the other hand, is the product of large-scale interlocking systems of extraction, production, distribution and consumption, and such systems gain their own momentum and dynamic. \citep[][p. 5]{moore1997primitivist}
\end{quote}
On this, we can readily see an abacus can be both a tool and exist in a relationship to human cognition.
A familiar case of this type of offloading is sending off your desired arithmetic function and the arguments, e.g. adding two numbers, to a digital calculator.
In this example, you do not add the numbers yourself.
You do three things: know which numbers you need, know which function you need, and know how to use the machine to offload what would otherwise be your cognitive labour.
This is step one: yes, it is AI.
Had the machine not existed, you would have had to perform the arithmetic yourself, and arithmetic is cognitive labour, therefore, yes.
The AI relationship could be any of the three in \autoref{tbl:1}, i.e. more information is needed to understand what is going on in a specific use-case of a calculator to perform addition.
Let us pick as the human in this relationship, a familiar one to us all, a child who knows the symbols for numbers and addition, and how to use the calculator, but does not yet know mental or otherwise addition.
The effect that offloading addition to the calculator has in this case is why we ban their use by children who have not yet learned basic numeracy, i.e. in this case using a calculator embodies a way to avoid learning how to add numbers from the rote learning of the addition of pairs of numbers under ten to the rules for larger numbers.
As a society then, we decide it is undesirable and mostly the 3rd column of step 2 in \autoref{tbl:1}: \textit{a}) displacement.
It is \textit{b}) harmful to the development of \textit{c}) the child's numeracy skills and solely knowing how to add using a calculator is not knowing how to add because problems will appear when, e.g. the numbers to be added are beyond the maximum number representable by the calculator, in such a case \textit{d}) obfuscation of somebody else's cognitive labour is likely to happen as the calculator on its own without somebody who knows addition is not \textit{e}) equivalent to a human.\footnote{In the case of adding numbers outside the maximum, a person who does know addition can use the calculator to perform parts of the sum or resort to paper or any other combination of the above.}

For a calculator generally, typically the human-in-the-loop, involvement of human oversight or data after the user input, \textit{f}) is either absent or minimal (e.g. to change the batteries) --- and relatedly, the input \textit{g}) is easily identifiable as such, i.e. what the user must have done to obtain \textit{h}) the desired output, in this case the result of the arithmetic operation requested, which is formally well-specified and verifiable.
Importantly, a different relationship to humans as calculators typically enjoy is not displacement AI, but likely enhancement AI: with calculators we lose none\footnote{Perhaps skills like long division atrophy, but we tolerate this.} of our extant mathematical skills as adults and enjoy a shorter time complexity on numerical operations (see column 2, \autoref{tbl:2}).
Not all technology has such a trajectory, as we shall see with specific cases of this in-depth in the next section.
% \autoref{sec:avscl}: \nameref{sec:avscl}.

% \setcounter{section}{1}
\section{Artefacts versus Cognitive Labour}\label{sec:avscl}
\begin{quote}
    There can be no doubt that the idea of ``intelligent'', ``thinking'' machines has captured the imagination of many people all over the world. \citep[][p. 295]{Saparina_1966}
\end{quote}
% Alarms, photos, and clothes}
% \begin{quote}
%     Initiation into a trade and becoming accepted as a skilled worker no longer takes place by way of institutions, or at least not those envisaged in such statements as `the skill has precendence over the machine'. With industrial capitalism the spasmodic evolution of machinnery keep cutting across the existing hierarchy of skills.
%
%     \citet[][p. 112]{guattari1984molecular}
%     % p. 112, guattari % https://mydesiringmachines.wordpress.com/wp-content/uploads/2014/08/felix-guattari-machine-and-structure.pdf
% \end{quote}
AI is unlike a tool like a saw used to cut wood where the person cutting also puts in labour, often more than the creator of the saw, to cut wood, in the sense that a saw is involved in the predominantly overt physical and cognitive labour of woodworking \citep{guest-inprep}.
% When you cut wood with a saw you experience the sociotechnical relationship as enhancement (row \textit{2a},  \autoref{tbl:1}) and beneficial (\textit{2b}) because:
% you learned or refined your woodworking skill (\textit{2c});
% even if the production of the saw before you purchased it all depended on a long pipeline from raw materials to saw, nothing about you using a saw obfuscates this pipeline as an aspect of the saw over and above any other technology or tool (\textit{2d});
% you likely could not have cut the wood without a saw, it exceeds human equivalence (\textit{2e})
% you alone did the cutting (\textit{2f});
% your labour of moving the saw and positioning where to cut the wood is apparent (\textit{2g});
% wood really was verifiably cut (\textit{2h}).
Not so for typical cases of contemporary AI, like chatbots --- in contrast:
You did not contribute other than as input, e.g. the so-called prompt, ultimately harming the chance to learn anything substantial \citep[e.g.][]{bastani2024generative, guest-inprep}.
It is never clear if the chatbot's results really match those desired, e.g. so-called hallucinations --- ``a misleading (and anthropomorphizing) term[, which has become mainstream, and thus] a major win for AI hype'' (\citealt{helfrich2024}, p. 700; \citealt{hicks2024chatgpt, bishop2021artificial}).
Due to their opaque nature, AI relationships hide labour that might be invoked not only like the saw to extract raw materials or design systems, but also in real-time as you use the artefact, e.g. you forget that sweatshop workers may be in real-time or the recent past refining the output of the chatbot \citep{crawford2021atlas, perrigo2023exclusive, placani2024}.
% AI users, in contrast, are customers much more like the person buying the end product of woodwork than carpenters themselves, if not in some sense more so as they are remain unaware of and are even tricked into thinking they performed the relevant labour (e.g. obfuscatory phrases like ``prompt engineer'' imply labour and skill).
% Being a customer or user not only requires no training, is not a skill, but also is a completely different labour relation masquerading as skill.
% This is reflected in the advertising deployed by AI companies who claim that their models allows one to ``talk to a computer'' and that they require no skills, e.g. programming, to do so.
% And this deskilling is not containable.
% Seeing students merely as customers of AI products, for example, deskills not only them, but their teachers: there is nothing to teach.
So much for a stark contrast between a tool-human sociotechnical relationship without AI and a relationship that involves modern chatbots, what about technologies that are more complex than a saw?

Below, three triplets are presented for exploring and exemplifying my definition of AI (from \autoref{tbl:1}):
\begin{description}
    \item[\ref{sec:acc} \nameref{sec:acc}]: all three artifacts (in \autoref{tbl:2}), which are prototypical computational aids or devices, diverge greatly on their need for direct human involvement, but nonetheless can share the same desired output (the result of a given calculation).
    \item[\ref{sec:acf} \nameref{sec:acf}]: these three examples (in \autoref{tbl:3}) demonstrate how divergent and different the cognitive labour and capacities (knocker-upper, the human alarm clock; human vision; seamstress/tailor) and their related artefacts are.
    \item[\ref{sec:lic} \nameref{sec:lic}]: these three (in \autoref{tbl:4}) implicate what is typical contemporary AI, very deep artificial neural networks trained on extremely large datasets \citep[a correlationist programme previously dubbed \textit{modern connectionism} in][]{guest_martin_2024} with the cognitive labour they claim to capture, of which all relationships are characterised as displacement.
\end{description}

\begin{table}[tbh]
% \begin{samepage}
\caption{Three examples (columns) of technosocial systems (e.g. user and alarm clock) paired to what came before (knocker-upper; column 1) or to a classical cognitive capacity (vision; column 2) or to a non-sweatshop version of similar skills and labour (seamstress/tailor; column 3) to demonstrate that their important properties (rows) can be teased apart and understood if we center humans in our analyses (recall \autoref{tbl:1}).
% In the table, each artifact (and related emdedding) is paired with its pre-existing cognitive labour: the alarm clock with what it replaced, the knocker-upper, a profession; the camera, which can be seen as the successor to many technosocial embeddings, including visual artists, but here we select its relation to vision; and the garment factory's relationship to the tailors who make bespoke clothes.
}
\input{table3}
% \end{samepage}
\label{tbl:3}
\end{table}

\subsection{Abacus, Calculator, Computer}\label{sec:acc}
Analysing these ancestral forms of computers --- which function as aids, in the case of the abacus, and as perhaps something more independent in the cases of the calculator and Turing-complete electromechanical and electrical digital computers --- as AI per \autoref{tbl:1} brings to light aspects that are central to understanding cognitive labour.
For the abacus versus mental arithmetic AI relationship (see column 1, \autoref{tbl:2}), we --- surprisingly or not depending on our familiarity with abacus use --- see a marked benefit to mental arithmetic and an unambiguous development of a new skill \citep[][]{wang2020review,lima2021cognitive, lu2023transfer,xie2024long}.
Importantly, the examples in \autoref{tbl:2} have been chosen inter alia because we can keep row \textit{h} constant: all three have the same desired output in these cases.
An interesting highlight is row \textit{g}) the abacus is completely unable to perform any arithmetic operation without the abacist.
If you forget how to manipulate the beads, the abacus is merely decorative.
This is very different to the electronic calculator (see column 2, \autoref{tbl:2}) where one need not be familiar with the operation of the calculator at all: you only need to recall the symbols that represent numbers and functions.
Not knowing how to use a calculator is nigh on impossible in the modern world, even without having ever used one, since the prerequisite knowledge is reading and button affordances; no specific training is needed.
To forget how to use a calculator is to suffer significant cognitive impairment beyond being rusty moving beads around; one would need to lose the ability to read numbers and press buttons --- highly unusual.
And the opposite, using a calculator can harm children's ability to accomplish basic numeracy, and so we proscribe its use in primary schools for this reason.
In the general case, once mental arithmetic is mastered the AI relationship between calculator and human is overall neutral without offering any new skills but also providing predictably speedy and verifiably correct arithmetic.
The thermostat is another great example of no deskilling on an individual level --- before that we could only sense using our bodies if things were too hot or too cold, and we still can.

In contrast to these positive and neutral AI relations is the original human-computer relationship (see column 3, \autoref{tbl:2}).
A human computer was a person --- often a woman in the previous century, but less gendered prior to that --- who performed calculations, worked with computing machinery, wrote programming software \citep[][]{grier2013computers, shetterly2016hidden}.
For example, during Britain's war effort against the Nazis:
\begin{quote}
    Arriving members of the [Women's Royal Naval Service] were given two weeks training in binary math, the teleprinter alphabet, sight-reading punched paper tapes, and the structure and workings of the Tunny and Colossus machines. [...]

    Machine work---and the theory and skills it required---was an integral component both intellectually and functionally of [the Second World War's] codebreaking process. It was not, as many assumed due to its feminized nature, deskilled drudge work. \citep[][pp. 39--40]{hicks2017programmed}
\end{quote}
The same holds for the human computers in the USA at the National Aeronautics and Space Administration \citep[NASA;][]{shetterly2016hidden}:
\begin{quote}
    Early on, when they said they wanted the capsule to come down at a certain place, they were trying to compute when it should start. I said, ``Let me do it. You tell me when you want it and where you want it to land, and I’ll do it backwards and tell you when to take off.'' That was my forte. [...]
    
    But when they went to computers, they called over and said, ``tell her to check and see if the computer trajectory they had calculated was correct.'' So I checked it and it was correct. \citep[Katherine Johnson, as quoted in][n.p.]{hodges2008she}
\end{quote}
And in both countries these women were written out of the historical record; their erasure facilitated by the rise of the electronic computer.
This general pattern of displacement of women by machines, which we have previously dubbed \textit{Pygmalion displacement} \citep[][]{erscoi_kleinherenbrink_guest_2023}, and that of people by machines generally, has had harmful effects on society \citep{agar2003government, sherwood1985engels, adler1990marx, Wendling2002}, like the permanent harm to the British computer industry in the 20th century \citep{hicks2017programmed}.
Importantly, ``digital computers were promoted as more efficient and less error-prone than humans at calculations. But in fact this comparison is not `like-for-like' since, for example, calculating ballistic trajectories is, when done by women, also open to ethical questioning.'' \citep[][p. 20]{erscoi_kleinherenbrink_guest_2023}

Centring the human cognitive component, as well as outlining the artefact in the relationship with the definition in \autoref{tbl:1}, teases out important differences between the presented relationships.
For example, all three pairs have the same desired output, which is the result of the calculation (row \textit{h}, \autoref{tbl:2}) the abacus requires all of human cognition to work,  (rows \textit{f} and \textit{g}), while the calculator really does take over arithmetic, and the digital computer can take over even more, assuming the programmer can code it: ``algorithms are always already made, maintained, and sustained by humans.'' \citep[][p. 52]{bucher2018}
This also underlines how easily significant cognitive labour can be obfuscated when we move from left to right in \autoref{tbl:2}: every use of the abacus has obvious manipulation effort while once a computer is programmed, software runs without any direct indication it was handcrafted (row \textit{d}).
In other words and relevant for our fields, ``given the Cartesian legacy of the cognitive sciences, computers are looked at with veneration as soon as they produce well-formed output \citep[][]{weizenbaum1976computer}'' (\citealt[][p. 312]{rasenberg2023reimagining}; \citealt{jucan2023, powell1970descartes}).
We should therefore be on high alert when others (or we) program computers to perform complex tasks, so as not to be taken in by this and misled into thinking something mystical --- something other than a machine obeying our formal instructions --- has occurred.

\subsection{Alarm Clock, Camera, Garment Factory}\label{sec:acf}
Moving away from arithmetical operations and  Turing-complete comparisons, to specific artefacts outside obviously computational devices: \autoref{tbl:3} depicts the AI relationships between alarm clocks, cameras, and garment factories and respective selected cognitive capacities.
An alarm clock is a simple device that we provide some basic inputs to (row \textit{g}, column \autoref{tbl:3}) for it to function as we desire, e.g. to ring at a specific time (row \textit{h}).
To the untrained eye, an alarm clock may appear non-cognitive, and yet depicted in the black \& white photo in column 1, \autoref{tbl:3}:
``Mrs Mary Smith wakes the dockers of Limehouse, London, with her peashooter in 1931. [...] She was a knocker-upper, a human alarm clock'' \citep[][n.p.]{JohnTopham1931}.
The alarm clock completely automates every aspect of her kind of profession, with the exception there is no guarantee the user will be awake (row \textit{e}, column \autoref{tbl:3}).
Unlike with a human alarm clock, an artefact cannot promise the desired output --- we all have experienced sleeping through loud noises or some other alarm malfunction, and on the flip-side children are often woken by their parents to enure they make it to school.
Additionally, the inclusion of alarm clock functionality in mobile phones means their use can be further generalised through the day as reminders, with the inputs staying the same as an old-fashioned alarm clock and the desired output always requiring human supervision, e.g. a reminder to take out the rubbish is merely a reminder and not a guarantee the bags are taken out.
All this is very familiar to those experiencing struggles with the cognitive capacity of executive function.

A slight tangent here is useful on the history of clocks, which is one of control of the users by the measurement of time, and not the other way round, which reflects and underlines the need to reorient all AI into human-centric focus.
Prior to clocks, human labour was governed by the natural passing of time sans measurement, e.g. waking with the sun.
Clocks are hegemonic tools, backbones of industrialism and capitalism:
\begin{quote}
The problem of the clock is, in general, similar to that of the machine. Mechanical time is valuable as a means of co-ordination of activities in a highly developed society, just as the machine is valuable as a means of reducing unnecessary labour to the minimum.
Both are valuable for the contribution they make to the smooth running of society, and should be used insofar as they assist [people] to co-operate efficiently and to eliminate monotonous toil and social confusion. But neither should be allowed to dominate [people's] lives as they do today. \citep[][p. 8]{woodcock1944tyranny}
\end{quote}
This aspect of machines,  when they measure and control us, is one to bear in mind, and one which we will return to time and again below.

The camera is perhaps a more understandable addition to \autoref{tbl:3} with its relationship to the human capacity of vision well-known.
The point here --- as with all the other relationships --- is not a mechanistic similarity but one based on functional correspondence or role, factual or perceived \citep[cf.][]{guest_martin_2024, guest2023logical, pittphilsci25289}.
Notable in our relationship with the camera is the fact it has largely deskilled nothing in the present for the typical user, using a camera or looking at its output, does not negatively impact our ability to visually perceive (rows\textit{ b} \& \textit{c}, \autoref{tbl:3}).
Photographs also serve, especially in the advent of mobile phone cameras, to help us recollect our memories, enhancing our own ability to think about the past (row \textit{a}).
Of course, the user, the photographer, must frame and perform automatic or manual adjustments to the lens, for any image or video to be of use (row \textit{g}).
However, this human contribution, which manifests as the user, is ever present in all our AI systems from calculators to alarm clocks to much more complex systems like digital computers.
The more infrequently discussed human-in-the-loop is humans other than the user, such as cases like digital computers (3rd column of \autoref{tbl:2}) which require many other people's labour, not just the user's \citep{kalluri2025computer, birhane2023hate, kalluri2023surveillance, birhane2024dark, birhane2023into, couldry2019, crawford2021atlas, placani2024}.
An infamous example of the human-in-the-loop technique is the orientalist Mechanical Turk, which toured from the late 18th to the mid 19th century, wherein a person hid in a cabinet under what appeared to be an automaton that played chess \citep{stephens2023mechanical}.
In fact, the person below the puppet controlled its movements, giving the human player sitting across it the impression that they were being beaten at chess by a clockwork machine.
This is also the namesake of Amazon's Mechanical Turk, a platform on which low paid workers toil to ``earn pennies or dollars doing tasks that computers cannot yet [and may never] easily do.'' \citep[][n.p.]{newman2019found}

The sociotechnical relationship between the garment factory, which comprises humans-in-the-loop, can be seen as a more harmful, and equally obfuscatory, variant of the Mechanical Turk.
Workers are treated badly in many cases, such as where sweatshop labour is used, the environmental impact of so many clothes is ignored, and the harmful chemicals are glossed over.
In the not so distant past, people owned fewer clothes and had them hand- and custom-made by seamstresses and tailors.
We readily have accepted this displacement of a relationship that involved a skilled adult making us a small selection of well-fitting clothes, which typically could last a whole lifetime, to a world in which we consume clothes relentlessly made often by workers, who can even be underage, and are often in harmful conditions \citep[cf.][]{dyerwitherford2019, perrigo2023exclusive, stephens2023mechanical, crawford2021atlas, Wendling2002, steinhoff_2021}.
Notwithstanding, bespoke tailoring is to this day understood to be superior, because it factually is, and so preferred by the rich and famous and indeed required by anybody outside the bounds of factory-made standardised clothing configurations and sizes (row \textit{e}).
Even the simpler skill (compared to sewing from scratch) of taking in or out clothes as our bodies change over time is abandoned (row \textit{c}) as a function of the garment factory which produces new clothes cheaply that we can buy instead (row \textit{e}).
We now turn to the cutting edge of contemporary AI.

\begin{table}[tbh]
\caption{Contemporary AI products and models, such as LLMs, image generators, and chatbots are framed such that they are in competition with, or seen as equivalent to, cognitive capacities like essay writing, creating artwork, and providing companionship.}
\label{tbl:4}
\input{table4}
\end{table}
\subsection{LLM, Image Generator, Chatbot}\label{sec:lic}
If we centre the cognitive labour when teasing apart these three AI relations, as \autoref{tbl:1} guides us to do, it falls into our lap how it is displacement, harmful, and deskilling in every case: large language model (LLMs) versus essay writing, image generator versus artist, and chatbot versus human companionship (rows \textit{a}--\textit{c}, \autoref{tbl:4}).
Importantly, as \citet[][p. 312]{rasenberg2023reimagining} explain, ``whereas in human-animal interaction there is ample evidence of reciprocal adaptation, here the adaptation is strikingly one-sided, with [such systems] essentially helpless, requiring care \citep{lipp2023caring} and forcing people to adapt to their constraints \citep{alavc2020talking, suchman2019demystifying}.'' 
We are expected somehow to learn what is presented as a non-skill, because `prompt engineering' is indeed not a skill, to enable us to use these opaque corporate-owned stochastic context-addressable systems.
% which are advertised as requiring no skills.
And, as turbo charged versions of the Mechanical Turk, contemporary AI models ``are built on massive exploitative `ghost labour'; crowdsourced and outsourced labour that follows the patterns of colonial relations \citep[e.g.,][]{bender2021}'' \citep[][p. 3]{mcquillan2024we}.

The false advertising --- there is no engineering in prompt engineering --- grows the longer the rhetoric around these systems are examined because, as we have seen many times so far, ``data can only do so much. [In any apparently successful application of AI t]he real work is carried out by the people on the ground'' \citep[][n.p.]{Cybernet31}.
While in the previous century, ``computing systems functioned due to vast arrays of human workers, expressed through workflow organization, operators’ actions, and software'' \citep[][p. 5]{hicks2017programmed}, in the present the obfuscated human-in-the-loop --- such as the sweatshop workers who guide and power LLMs and other such systems, or us ourselves whose data is stolen without our knowledge --- is not respected \citep{o2016weapons,erscoi_kleinherenbrink_guest_2023,mcquillan2022resisting,Birhane_Guest_2021,guest2024teaching}.
Even the low bar human computers' treatment has set is not met by modern AI's dehumanisation and theft of labour \citep[e.g.][]{van2024artificial, Brennan2025, crawford2021atlas, perrigo2023exclusive, goetze2024ai, rhee_2024}.
In the time of software updates, the obfuscation of programmers' labour with seamless updates grows, while in the past one would physically go get artifacts fixed (or indeed fix them ourselves).

The systems in \autoref{tbl:4}, embody an obfuscation of labour  so complete the user believes the machine thinks for itself \citep{polo2024pensamento}.
In reality, exploited sweatshop workers in the Global South who perform the human-in-the-loop role do a lot of what we consider automated by AI \citep{perrigo2023exclusive, bender2021, mcquillan2024we, BAINBRIDGE1983775, Brennan2025, strauch2017ironies,crawford2021atlas}.
\begin{quote}
    % A commodity is therefore a mysterious thing, simply because in it the social character of [people]'s labour appears to them as an objective character stamped upon the product of that labour[.]
    % ; because the relation of the producers to the sum total of their own labour is presented to them as a social relation, existing not between themselves, but between the products of their labour.
    % This is the reason why the products of labour become commodities, social things whose qualities are at the same time perceptible and imperceptible by the senses.
    % In the same way the light from an object is perceived by us not as the subjective excitation of our optic nerve, but as the objective form of something outside the eye itself.
    % But, in the act of seeing, there is at all events, an actual passage of light from one thing to another, from the external object to the eye. There is a physical relation between physical things.
    % But it is different with commodities. [...]
    % There, the existence of the things \textit{quâ} commodities, and the value relation between the products of labour which stamps them as commodities, have absolutely no connection with their physical properties and with the material relations arising therefrom.
    There it is a definite social relation between [people], that assumes, in their eyes, the fantastic form of a relation between things.
    In order, therefore, to find an analogy, we must have recourse to the mist-enveloped regions of the religious world. In that world the productions of the human brain appear as independent beings endowed with life, and entering into relation both with one another and the human race.
    So it is in the world of commodities with the products of [people]'s hands.
    % This I call the Fetishism which attaches itself to the products of labour, so soon as they are produced as commodities, and which is therefore inseparable from the production of commodities.
\citep{marx1867fetishism}
\end{quote}
The reasoning problems become evermore severe in our misunderstandings of these most modern machines \citep{guest_martin_2024, guest2023logical, pittphilsci25289, pittphilsci24834}.
As Taina \citet[][p. 50]{bucher2018} explains: ``When a machine runs smoothly, nobody pays much attention, and the actors and work required to make it run smoothly disappear from view \citep[][]{latour1999pandora}.''
A next step in this devolution and devaluation of cognitive labour is that now the user too, deskilled and displaced, also disappears from view. 
What voice does the human, now reduced to \textit{only} a user, have if their essays (column 1, \autoref{tbl:4}), their visual expression (column 2) are just the copy-pasted output of a device that performs patchwork plagiarism?
What human connection do they have when their friends and romantic partners have been replaced with inanimate objects (column 3)?

The technopositive argument used to be that violence in video games was not indicative or causative of interpersonal violence because video game characters are virtual and users know the interaction is not in any way equivalent to that with other people outside the game.
If that logic still holds then users will gain no, or very few, positive effects if they are lonely and need companionship (column 3, \autoref{tbl:4}).
If that logic does \textit{not} hold, and people see these entities as possessing minds or as people, much more has unravelled.
Indeed, for certain users, the psychological damage caused by only or mostly interacting with entities known to be designed to trick the user into believing they are people is immense (\cite{turkle2005second, turkle2006relational, erscoi_kleinherenbrink_guest_2023, weizenbaum1976computer, PeopleAr24, weizenbaum1966eliza, jucan2023, dillon2020eliza, TheyAske37, Strengers2024, placani2024}; even the companies involved accept this potential harm: \cite{phang2025investigating}).

As mentioned, past worries about sociotechnical relations were along these lines:
\begin{quote}
    In a sane and free society such an arbitrary domination of [humanity's] functions by either clock or machine would obviously be out of the question. The domination of [humanity] by the creation of [humanity] is even more ridiculous than the domination of [humanity] by [humanity]. \citep[][p. 8]{woodcock1944tyranny}
\end{quote}
But in the present, and as \citet[][]{marx1867fetishism} notes above, there is the domination of humans by other humans via religious-, conspiratorial-, or cult-like logic \citep{guest_martin_2024, guest2023logical, SiliconV44, PeopleAr24, WhySilic59, hao2025empire, heffernan2025religious} and through these machines, algorithms, models.
Humans' expressions and social relations are not even mediated through technology, like when using a texting application to communicate with another person, but constitute technology as such.
Technology, in this scheme, controlled by a private company (as all examples in \autoref{tbl:4}) produces our so-called self-expressions.
These relations are no longer captured by already flawed metaphors like the echo chamber --- the echo has gone, the chamber is devoid of people: we neither shout nor are heard.
We abandon our voice, forget how to use it, and forfeit what makes humans special in all columns of \autoref{tbl:4}.
In centring cognition as the analytical tool in \autoref{tbl:1} ensures, we are forced to look the human-in-the-loop in the eyes and recognise that these relations are harmful to people.

% So where to from here?
% In the table, each artifact (and related emdedding) is paired with its pre-existing cognitive labour: the alarm clock with what it replaced, the knocker-upper, a profession; the camera, which can be seen as the successor to many technosocial embeddings, including visual artists, but here we select its relation to vision; and the garment factory's relationship to the tailors who make bespoke clothes.

% Networks of labor and expertise extend into the systems themselves, constructing the social and technological bedrock on which all computing projects rest. '' 

% First, we will examine the relationships between some aspects of human cognitive labour (mental arithmetic, pen-and-paper arithmetic, and human computers) and tools (abacus) and technologies (calculators and digital computers).
% After that we will explore more diverse human capacities versus technologies that relate to them.

% \subsection{Abacus, calculator, computer}

% \subsection{Alarm clock, camera, clothes factory}

% These phenomena sound minimally undesirable and maximally harmful.
% In the rest of this section, I go over three examples of AI relationships that do not have labour obfuscation to this maximal extent --- and I end on LLMs which do.

% \subsection{Notes}

\FloatBarrier

\section{Machine Hauntology \& Spectral Technology}
Ignoring the ghosts in our machines harms us and our understanding of all such systems, from scientific models \citep[][]{morgan1999models, guest2021computational, van2024reclaiming, powell1970descartes} to chatbots \citep[][]{turkle2006relational, dillon2020eliza, jucan2023, erscoi_kleinherenbrink_guest_2023}.
% ``algorithms are always already made, maintained, and sustained by humans.'' \citep[][p. 52]{bucher2018}
% In the context of cognitive science, \citet{Ryle1949, tanney2009ryle, van2008tractable, egan2025deflating} amongst others provide sensible starting points for reconsidering and formally structuring our theories and models for cognition, which maintain the machine and the computational but without obfuscation and the other typical reasoning traps \citep{guest2023logical, pittphilsci24834, pittphilsci25289}.
To truly centre the human in AI, we must admit the human's direct and inherent centrality, and such an admission can be facilitated by the radical redefining of AI in \autoref{tbl:1} and the examples unpacked in Tables \ref{tbl:2}--\ref{tbl:4}.
And so if HCAI wishes to uphold and enact its human-centred-ness, as a field or perspective, it must then aspire to properly address  each of the four points that are core to its current formulation  (listed in \autoref{sec:intro}, repeated here), then it must:
\begin{enumerate}
    \item with respect to supporting or enhancing human skills, both without displacement of said skills and without violation of fundamental human rights, recognise and \textbf{act when displacement AI relationships take place};

    \item with respect to imbuing systems with various so-called human-aligned values, including those of explainability and transparency, realise that \textbf{human-aligned values can only exist in systems where we actively uphold those values}, they do not come for free, and are not formally guaranteed, but are constantly negotiated through sociotechnical struggles;
    
    \item with respect to focussing on human behaviour as a benchmark, i.e. the idea of human-like or -level performance, steer clear of correlationist, such as naive computationalist or modern connectionist, stances and roundly \textbf{reject benchmarks as meaningful};

    \item with respect to implicating behavioural, or otherwise psychological, methods in the study of these systems as if on equal epistemic footing with humans, take heed of serious warnings about our collective scientific reasoning as \textbf{correlations are red herrings in the search for theoretical understanding}.
\end{enumerate}
To do this, as I have demonstrated, the definition proposed in \autoref{tbl:1} changes how we see artefacts, freeing us to view many more artefact-human relations as having human cognition at their centres.
% Does it mean everything technological is AI?
% Likely yes.
% Are keyboards AI? Yes, they replace the cognitive skill of handwriting with the pressing of keys or of a location on a touchscreen that simulates a physical keyboard.
% Are (non-driverless) cars AI?
% Yes, for example, if we take a vehicle as replacing a horse and rider with a car and driver.
% Both the cognition of the rider and of the horse are replaced by the driver and the car.
% Importantly, the horse was not only contributing power, but cognition with respect to computing the path from A to B at the micro and medio scales, e.g. where to place hooves, how to move towards a certain direction given by the rider, and how to translate commands into a certain gait.
% Are books, or the written word generally, AI?
% Yes, if the alternative is memorising what would have been written.
% Is a calculator AI? Hopefully obviously so under this frame.
% For example and as mentioned above, our scientific models and simulations can be analysed as AI as they enable some automation of conclusions from our theoretical and other commitments \citep{guest2021computational, Morrison_Morgan_1999, van2024reclaiming}.
% Is a table AI? If the alternative is holding items, then yes.
% This might be pushing it, but it demonstrates that 
For the first part of my new proposed definition, if cognitive labour appears to be  outsourced to a greater or lesser extent to an inanimate object, we can call this relationship between technology and cognition: AI.
For the second step, even more analysis is needed wherein we need to discern and evaluate the relationship (recall rows in \autoref{tbl:1}) between humans and a given use of the artefact under question (recall Tables \ref{tbl:2}--\ref{tbl:4}).
%
%
% So we deflated AI by saying anything that can be framed as performing human labour is AI.
% To evaluate if something is outsourced human, or cognitive generally, labour human cognitive labour is needed; something hopefully not lost on the reader.
% Importantly, however, this is not the full story I want to tell.
% Deflation of AI as a phrase without reinflation into something useful as a critical lens is pointless.
% I would like to say what sort of sociotechnical relationship we are dealing with: enhancement, replacement, or displacement; as shown in \autoref{tbl:1}. 
% To answer the section's titular question.
%
%
% rework the following into subsections and using Table 2
%
% Why is such a definition useful, and what purpose does it serve?
Deflating allows us to do things that chasing after what AI currently is per fads of the technology sector does not --- specifically:
\begin{enumerate}[label=\alph*.]
    \item We can \textbf{centre human cognition}, and therefore we can centre the study thereof, cognitive science, as a relevant discipline to understand purported cases of artificial cognition \citep[e.g.][]{van2024reclaiming}.
    Under computationalism, which the possibility of engineering cognition implicates, correlations and benchmarks are not relevant \citep[e.g.][]{pittphilsci25289, pittphilsci24834}.
    This allows us to in principle reject any argument that uses behavioural or neuroimaging correlation to argue for human-likeness of an artifact.
    
    \item
    % It frees us from terms that are vacuous on deeper examination as playing a role in defining AI.
    We can \textbf{reject AI hype}, anthropomorphism, mysterianism about known mechanism,\footnote{The idea that engineered mechanisms, like matrix multiplication or other operations in artificial neural networks are somehow uniquely unknown or unknowable properties of these systems.} exaggeration, or fads, removing this rhetoric from being relevant as to what counts as AI, which is a typical frame with many definitions of AI, often to push products \citep[e.g.][]{forbes2025improve, duarte2024editorial}. The same goes for claims about neural or biological plausibility or inspiration; these have no useful coherent definitions \citep[e.g.][]{guest_martin_2024, guest2023logical}.

    \item We can \textbf{consider AI in general abstracted terms} without requiring specific reference to current advances in AI, e.g. chess-playing algorithms are AI, regardless of whether systems are cutting edge or not \citep[sometimes called the AI effect;][]{mccorduck2004machines}. And so we can easily reject that \textit{only} artificial neural nets or large language models or  generative AI are AI, especially when until recently GOFAI (good old-fashioned AI, also known as symbolic) was canonically AI, hence the name \citep[][]{guest-inprep}.

    \item Relatedly, we can \textbf{grant AI  a (pre)history}, allowing us to include the Antikythera mechanism \citep[][]{freeth2021model}, astrolabes, sextants, abacuses, and more in the timeline of AI \citep[e.g.][]{erscoi_kleinherenbrink_guest_2023, mayor2018gods}.
    We can uncondense time --- allowing us to slow down and giving us back our history --- which is centrally relevant for understanding our present or possible futures \citep{Hamilton_1998, stengers2018another}.
    As mentioned, this is something of a phobia, notably:
    \begin{quote}
    In English, the use of the word cybernetics raises no
    difficulties. Frenchmen with sufficient curiosity, however,
    were surprised to find it in Littre and Larousse; and the forgotten writings of Ampere were exhumed.
    When someone
    eventually turned up the new term in Plato, some of the
    experts rose in horror, declaring that kybernitiki should on
    no account be translated 'cybernetics'. \citep{guilbaud1960cybernetics}
   \end{quote}  
\end{enumerate}

Taken together, these properties and by-products of \autoref{tbl:1} allow us to perform transcendental as well as  immanent analyses of AI, such that we can  pick out more than artificial neural networks, or  specifically large language models, or such that we can perform analyses outside the tired contrasts of GOFAI or symbolic AI versus connectionist AI \citep{guest_martin_2024}.
No more mystification is possible because clarity and simplicity of definitions is within reach, and because these models are now correctly positioned on a historical timeline, and subject to scientific investigation outside the correlationist dogma \citep{guest_martin_2024, pittphilsci24834, pittphilsci25289}.
We do not need to passively be led astray, and in circles, by the technology sector any more, which ``is mostly concerned with building profitable artifacts  and is unconcerned with abstract definitions of intelligence.'' \citep[][p. 4]{Heffernan2019}

Instead we can focus on: finding persistent themes --- e.g. Pygmalion displacement: the displacement of women by machines and algorithms, a process that involves dehumanising the women while humanising the technologies \citep[][]{erscoi_kleinherenbrink_guest_2023} --- through teasing out properties of the sociotechnical relationship; avoiding hype as we centre human cognition%
% , the ghost, without assuming the machine does any work
; and using example artifacts from the distant and near past to help cut through exaggerated claims. 
Importantly, we can focus on the harms of the sociotechnical relationship on a case-by-case basis, while also learning from the past without getting bogged down by whether a specific system is, e.g. generative AI versus a convolutional neural network, which is discriminative AI \citep{mitchell1997machine, jebara2004generative, ng2001discriminative, efron1975efficiency, xue2008comment, guest-inprep}.
We sidestep  being tricked into using such formal terminology wrongly in service of the technology industry's spin game, which coopts and distorts formal terms to cause confusion and hype \citep[][]{guest-inprep, duarte2024editorial, helfrich2024}.
This kind of protection from misinformation is especially important as companies largely deploy closed source models, but even in the case of open source code often provide misleading or otherwise lacking documentation, limiting scientific investigation
% to black-box analyses
\citep[][]{liesenfeld2023opening, barlas2021see, ojewale2025towards, mirowski2023evolution, widder2024open, SamAltma61, birhane2023hate, kalluri2025computer, birhane2024dark}.
% Next, we walk through multiple examples of AI relationships to demonstrate the above honing of our critical thinking.
% Does this deflated meaning of AI cause harm?
% It might appear deflating AI to be potenaitally anything that can be in a technosocial relationship with cognitive labour can make us lose our ability pick out hamrful or modern AI.
% I propose this is not the case as AI is already used with many meanings (van Rooij Table 1), and thias may in fact force more specific terms such as LLM, ANN, statistics, and so on, which are (more) concrete terms for the systems we wish to critique and do not smuggle in (excep tin the case of LLM and ANN) terms that confuse or assert that intelligence really can be artificial.

% The (extreme) obfuscation of cognitive labour by technology --- a defining factor of what I dub \textit{displacement AI} (recall \autoref{tbl:1}) --- has been noticed by others \citep{pfaffenberger1988fetishised, BAINBRIDGE1983775, guattari1984molecular, erscoi_kleinherenbrink_guest_2023}.

Relatedly, the cyclic reasoning of treating machines as if we have decided already that they are human-like (points 3 \& 4 above) is evident at the birth of AI as a field.
As Teresa \citet{heffernan2024imitation}, explains:
\begin{quote}
Turing speculated that by the end of the century the ``use of words...will have altered so much that one will be able to speak of machines thinking without expecting to be contradicted'' (\citeyear[][p. 442]{turing_1950}), and many of today's AI researchers have, following Turing's lead, altered the meaning of words like --- ``reading,'' ``intuiting,'' ``feeling,'' ``dreaming,'' and ``creating'' --- to accommodate machine logic.

\end{quote}
And the same for other so-called founding fathers of AI, who when coining `Artificial Intelligence' claimed ``that every aspect of learning or any other feature of intelligence can in principle be so precisely described that a machine can be made to simulate it'' \citep[][p. 2]{mccarthy1955}.
This foundational document of AI, which was a proposal for a two month long summer conference, shows all the hallmarks we wish to avoid.
As described by \citet[][p. 47]{pettit2024}: ``The starting point for this multidisciplinary effort was the coupling of `natural and artificial intelligence,' although, as the Dartmouth proposal made clear, the distinction between the two was uninteresting for many.
These two realms constituted a common problem.''
In a now classic act of condensing time \citep[recall][]{stengers2018another, Hamilton_1998}, \citet{mccarthy1955} infamously proposed that this project of `solving' intelligence would take a single season.
Not in this AI summer, and not in any, will mathematically impossible displacements of humans pass muster \citep{van2024reclaiming, rich2021hard}.

More and more automation, much like a longer and longer road, does not imply that along the way something magical starts to happen merely as a by-product of distance between cause and effect or origin and destination.
Nothing qualitatively different than creating a series of pathways between locations is underway if we keep extending the road.
This being said, what does qualitatively shift is the a posteriori tractability of the emergent possible pathways if roads intersect and branch out.
My specific route to work is tractable and manageable by me, but the minutiae of the route are likely impossible to guess without further information the longer the distance is as the road network, and my own preference for stops along the way, for example, together provide infinite possibilities for which route I \textit{could} take.
There is no magic here --- other than metaphorically used perhaps for the awe human cognition should inspire --- the proverbial ghost in the machine is the literal human-in-the-loop \citep{pittphilsci24834}. 
The route is tractable to me because I picked it, while also being intractable to guess given a large enough search space for somebody else to copy.

This same metatheorising could be at play when we wrongly consider that engineered systems even more complex than roads as having qualitative shifts the more expansive the automation is --- consider an abacus versus a calculator versus a full-blown digital computer (as in \autoref{tbl:2}).
The only qualitative shift is in their formal complexity which is not a function of a quantitative aspect of a system.
The digital computer is closer to literal Turing-completeness, but no slight of hand is at play: we know computers are made-up of smaller components such as flip-flops and logic gates and it is not their \textit{quantity} that causes Turing-completeness.
Amusingly, although perhaps also depressingly for the state of computational literacy, LLMs are most likely less computationally expressive than the digital computers on which they run.
What is the case is the road, the conceptual and actual distance, from bare metal and silicon hardware to the software is longer.
LLMs appear above the physical computer on which they run, seemingly ensnaring us to assume such an extension of matter touches on the cognitive.
But like a longer road nothing has really changed in the engineered system: just more of the same.
This is unlike the biological and cognitive spheres.
And while unintuitive for many, the acceptance of this difference is the way forwards under computationalism.

I do not say the above  to motivate rejections of mainstream computationalism, but if the naive conception furthers mystifying basic computational processes and engineering then it surely must, we must make it, give way to more mature ways of thought that take into account what it is we commit to \citep[cf. basic versus naive versus non-naive computationalisms,][]{pittphilsci24834}.
In other words, just because the destination along the road is currently invisible to the naked eye from the origin does not by any stretch of the imagination imply something magical will get us there.
What gets us to the destination --- any goal --- is our pre-existing cognition whether the end is currently visible to the naked eye or not.
When it comes to the part of getting there that is automated, from car to calculator, the systems' parts are known and just hum along similarly regardless of whether the distance is 50 meters or 50 thousand or whether the digits are in the tens or the trillions.

Those who dogmatically insist that benchmarks have something to say about the humanity of machines assume that a behavioural match is informative, and that the correlationist programme can deliver evidence for their  assumption \citep[][]{guest2021computational, guest_martin_2024}.
But cyclic reasoning collapses here under its own weight.
There are infinite correlations to draw between an item and its reflection, but nothing in the, AI or traditional, mirror is the thing itself \citep{vallor2024ai}.
Such
\begin{quote}
    dogmatists are lazy-bones. They refuse to undertake any painstaking study of concrete things, they regard general truths as emerging out of the void, they turn them into purely abstract unfathomable formulas, and thereby completely deny and reverse the normal sequence by which [humanity] comes to know truth. \citep{tse1953contradiction}
\end{quote}
Truths, correlational, or otherwise do not emerge from the data.
They are the products of our cognition, and of our interactions with the world.
It is perhaps an uncomfortable truth that there are parts of cognition, like the human practice of science, that   cannot be automated \citep{rich2021hard, van2008tractable, Ryle1949, tanney2009ryle}.

% Neural network as doctor
% ``What makes an AI system seem trustworthy, and is that trust well-placed?''
% It mirrors back our own cognition because it is our cognition.
% In the cases where such systems, like clocks, accurately capture something about the world, like the time of day, it is because we understand something about the way time works.
% If we do not understand how something works, like how doctors diagnose, how humans adapt to novel siatuations, etc, then the AI just captures what we know about the world 

There are no Rube Goldberg-like perpetual motion-like human-in-the-loop-free machines.
There \textit{never} will be.
We might crave such a machine, but the only entities we know that are self-sustaining, autopoietic, is everything \textit{but} machines.
The only person who can decide to turn a computer off and then on again is exactly that: a person \citep{polo2024pensamento}.
To effectively avoid correlationism, obfuscation of cognition in AI, we must reject the current mainstream view:
% , described by \citet{Ryle1949} more than half a century ago, that ``the truth of
% mechanism is entailed by the truth of [the] theory of scientific research
% method in psychology'' (p. 301), i.e. 
no amount of high scores on benchmarks, or any other correlationary evidence, can ever pile up high enough to graduate to a causal claim.
% \citep[][p. 301]{Ryle1949} 
To de-fetishise AI, we must accept that AI is any relationship between technology, tools, models, machines and humans where it appears as if some cognitive labour is offloaded onto such artifacts, and furthermore we must accept that such a relationship requires methodical teasing apart to obviate the centrality of the human \citep[][]{morris2017after, braune2020fetish, mota2024fetichismo, pfaffenberger1988fetishised, bernardi_2024}.
We cannot rid the machine from its ghost.
But we can rid the concept of human from ghostliness --- the human ``need not be degraded to a machine by being denied to be a ghost in a
machine.'' \citep[][p. 301]{Ryle1949}

% about fetish 

%% file: table1.tex
\definecolor{paledogwood}{HTML}{D8BDBE}
\definecolor{lavendar}{HTML}{BD90E3}
\definecolor{indianred}{HTML}{cc575f}

\definecolor{mimipink}{HTML}{ffd6ff}
\definecolor{periwinkle}{HTML}{bbd0ff}
\definecolor{eggplant}{HTML}{533B4D}
\definecolor{champagne}{HTML}{FAE3C6}
\definecolor{hotpink}{HTML}{F564A9}
\definecolor{murray}{HTML}{990B4F}
\definecolor{mossgreen}{HTML}{87986A}

\newcommand{\HY}{\hyphenpenalty=0\exhyphenpenalty=0}

\centering

\begin{tblr}{
cells={font=\sffamily, valign=m},
colspec = {X[l]X[l]X[j]X[l]},
row{2} = {1.5\baselineskip},
row{3} = {6.75\baselineskip},
cell{2}{2-4} = {fg=white, font=\sc\sffamily, halign=c},
% column{4}={mossgreen!10!white, 103pt},
% column{3}={mossgreen!40!periwinkle!55!white, 10pt, fg=black},
% column{2}={periwinkle!15!white, 103pt},
column{4}={mossgreen!10!white, 37.3mm},
column{3}={mossgreen!40!periwinkle!55!white, 4mm, fg=black},
column{2}={periwinkle!15!white, 37.3mm},
column{1}={leftsep=0pt},
cell{3}{1}={valign=h},
cell{2}{4}={mossgreen!80!white},
cell{2}{2}={periwinkle!90!blue},
row{1}={bg=white, 2\baselineskip},
row{2-3}={rowsep=2pt},
}

\textit{1}) \textbf{Discern relationship:} & \SetCell[r=1,c=3]{l} Does the artefact relate to cognitive labour?&& \\

\textit{a}) \textit{entities involved} &Machine 
&
\SetCell[r=2]{c} AI
% \multirow{ 2}{*}{\textcolor{white}{\textbf{AI}}}
% \multirow{ 2}{*}{AI}
&Human\\
\textit{b}) \textit{potential terms} &
algorithm, artificial, automation, benchmark, discriminative, computer, engineering, functional, generative, mechanism, 
%mechanistic,
% large language model, model, neural network, 
% statistics,
system, technology, tool
&
% \multirow{ 2}{*}{\textcolor{white}{\textbf{AI}}}
&
ability, behaviour, capacity, cognition, 
% explain,
intelligence, labour, learn, organism, professional, psychology, reason,
% rights,
skill, task, thought, train
% , values
\\
\end{tblr}

\vspace{10pt}

\begin{tblr}{
cells={font=\sffamily, valign=m},
colspec = {X[l]X[j]X[j]X[j]},
column{1}={leftsep=0pt},
cell{even}{2} = {paledogwood!30!white},
cell{odd}{2} = {paledogwood!10!white},
cell{2}{2}={paledogwood!90!black},
cell{even}{3} = {lavendar!30!white},
cell{odd}{3} = {lavendar!10!white},
cell{2}{3}={lavendar},
cell{even}{4} = {indianred!30!white},
cell{odd}{4} = {indianred!10!white},
cell{2}{4}={indianred},
rows={1.5\baselineskip},
% row{2} = {1.5\baselineskip},
row{1}={bg=white, 2\baselineskip},
cell{2}{2-4} = {fg=white, font=\sc\sffamily, halign=c},
% column{2-4}={72pt},
column{2-4}={26.2mm},
row{1}={bg=white, 2\baselineskip},
row{2-Z}={rowsep=2pt},
% column{1}={90pt}
% column{1}={2.75cm},
% , valign=t},
% column{2}={4.5cm},
% % , valign=t},
% column{3}={4.5cm, valign=t},
% column{4}={4.5cm, valign=t},
}
 \textit{2}) \textbf{Characterise relationship:} & \SetCell[r=1,c=3]{l}How does the artefact relate to cognitive labour? &&\\
 % \textit{a}) \textit{label} & Enhancement & Replacement& Displacement \\
 % \textit{b}) \textit{valence} & beneficial & neutral & harmful \\
 % \textit{c}) \textit{effect on cognition} & reskilling & unaffected & deskilling \\
 % \textit{d}) \textit{labour obfuscation} & minor/none & unlikely & maximal \\
 % \textit{e}) \textit{human equivalence}& different & worse or same & no \\
 % \textit{f}) \textit{human\-/in\-/the\-/loop} & possible & rare & common \\
 % % b)& \textit{deskilling} & maybe & no & yes \\
 % \textit{g}) \textit{human input} & transparent & transparent & opaque \\
 % \textit{h}) \textit{desired output} & formal & specified & unspecified \\ 
 % % \textit{i}) \textit{human outperformance} & no & depends & no \\
 % % $\vdots$ & $\vdots$ & $\vdots$ & $\vdots$ \\
\textit{a}) \textit{label} & Replacement & Enhancement & Displacement \\
\textit{b}) \textit{valence} & neutral & beneficial & harmful \\
\textit{c}) \textit{effect on cognition} & unaffected & reskilling & deskilling \\
\textit{d}) \textit{labour obfuscation} & minimal & possible & maximal \\
\textit{e}) \textit{human equivalence} & worse or same & different & no \\
\textit{f}) \textit{human-in-the-loop} & rare & possible & common \\
\textit{g}) \textit{human input} & transparent & transparent & opaque \\
\textit{h}) \textit{desired output} & specified & formal & unspecified 

\end{tblr}

%% file: table2.tex
\definecolor{paledogwood}{HTML}{D8BDBE}
\definecolor{lavendar}{HTML}{BD90E3}
\definecolor{indianred}{HTML}{cc575f}
\definecolor{mintgreen}{HTML}{BEE3DB}
\definecolor{cambridgeblue}{HTML}{89B0AE}

\definecolor{mimipink}{HTML}{ffd6ff}
\definecolor{periwinkle}{HTML}{bbd0ff}
\definecolor{eggplant}{HTML}{533B4D}
\definecolor{champagne}{HTML}{FAE3C6}
\definecolor{hotpink}{HTML}{F564A9}
\definecolor{murray}{HTML}{990B4F}
\definecolor{mossgreen}{HTML}{87986A}
\definecolor{verdigris}{HTML}{18A3A5}

 \centering
 \sffamily%\footnotesize
    \begin{tblr}{
        colspec={X[l]X[l]X[l]X[l]X[l]},
        cells={valign=m},
        row{1}={white, halign=c},
        cell{1}{1} = {font=\itshape\bfseries},
        column{1} = {halign=l, leftsep=0pt},
        column{2-Z}={31mm},
        % column{2}={30mm},
        % column{3}={31mm},
        % column{4}={32mm},
        cell{1}{2-Z} = {font=\sc\sffamily, fg=verdigris!50!black},
        row{4-Z}= {2\baselineskip},
        cell{odd}{2} = {mintgreen!15!white},
        cell{odd}{3} = {mintgreen!50!cambridgeblue!15!white},
        cell{odd}{4} = {mintgreen!15!white},
        % row{2-3}={bg=white, halign=c},
        row{2-3}={halign=c},
        row{2}={2.5cm},
        row{3}={3.5cm},
        % row{Z}={halign=c},
        cell{1}{2}={bg=mintgreen, halign=c},
        cell{1}{3}={bg=mintgreen!50!cambridgeblue, halign=c},
        cell{1}{4}={bg=mintgreen, halign=c},
    }
 \makecell[l]{artefact \\ versus \\ cognitive labour} &
 \makecell[c]{Abacus \\ versus \\ Mental Arithmetic} &
 \makecell[c]{Calculator \\ versus \\ Human} &
 \makecell[c]{Digital Computer \\ versus \\ Human Computer} \\
 & \includegraphics[height=18mm, valign=b]{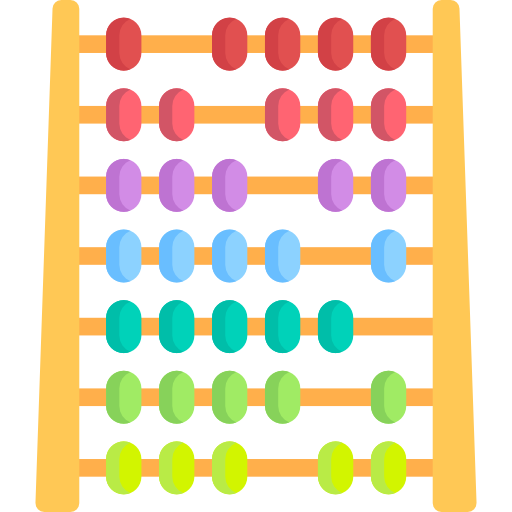} & 
 \includegraphics[height=18mm, valign=b]{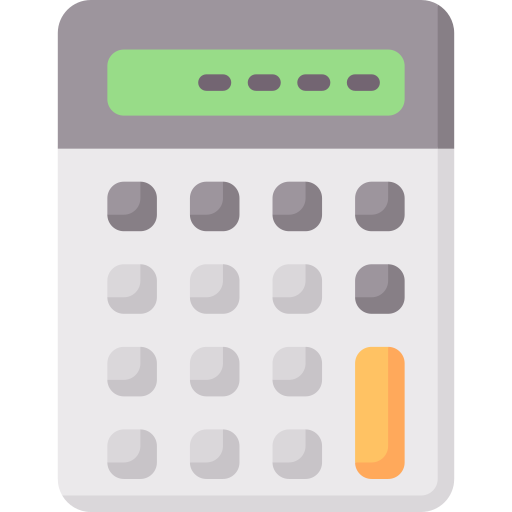} &
 \includegraphics[height=18mm, valign=b]{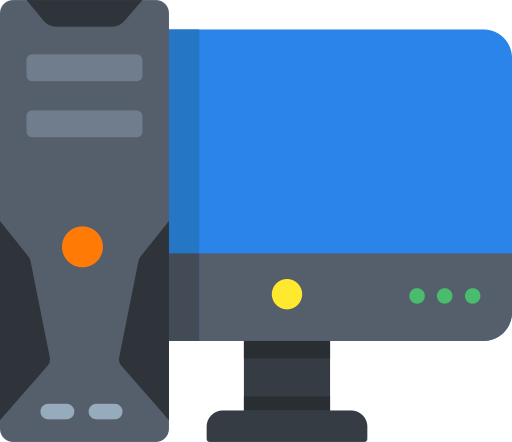} \\
& \includegraphics[width=30mm,valign=m]{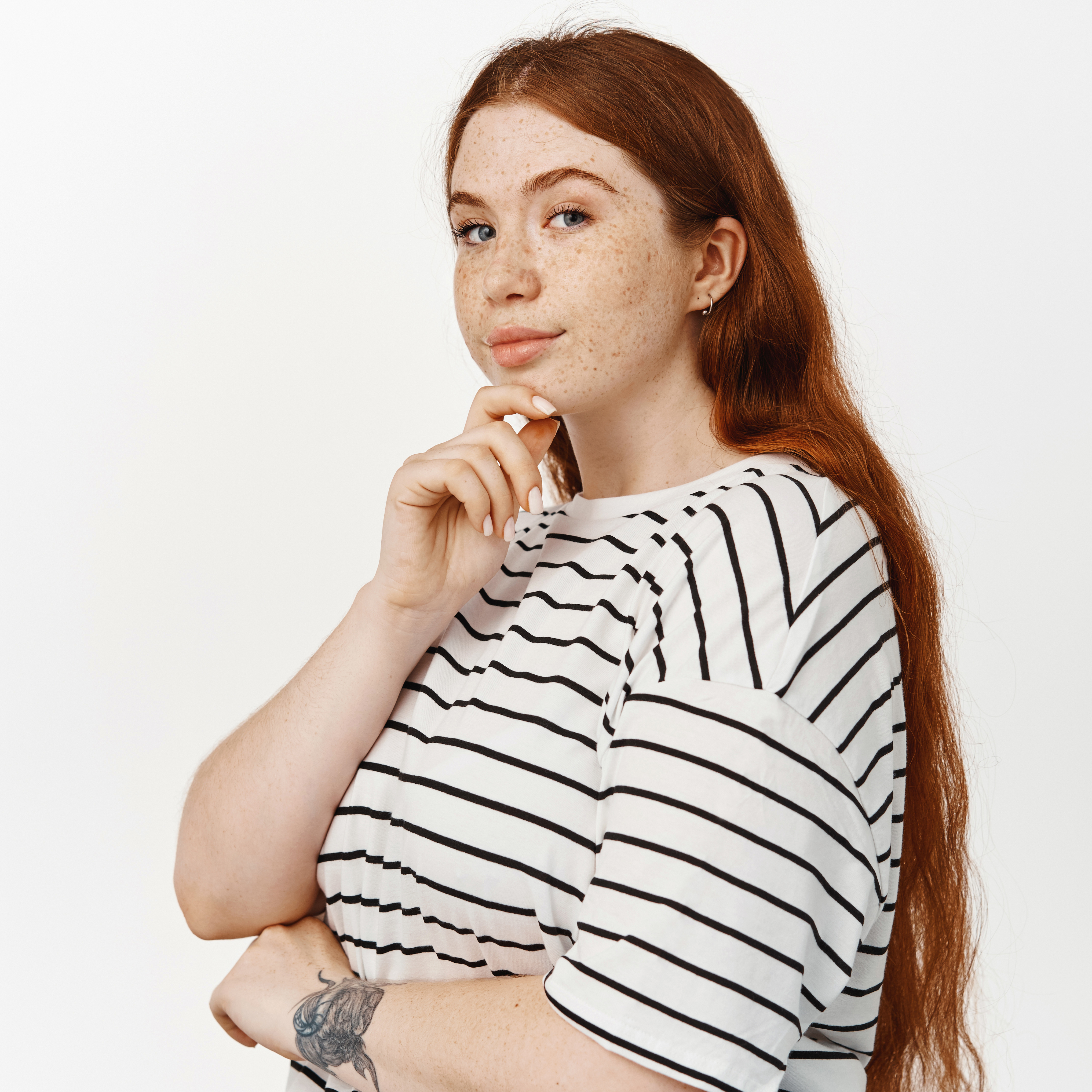} & % https://www.computerhistory.org/collections/catalog/102741216
\includegraphics[width=30mm,valign=m]{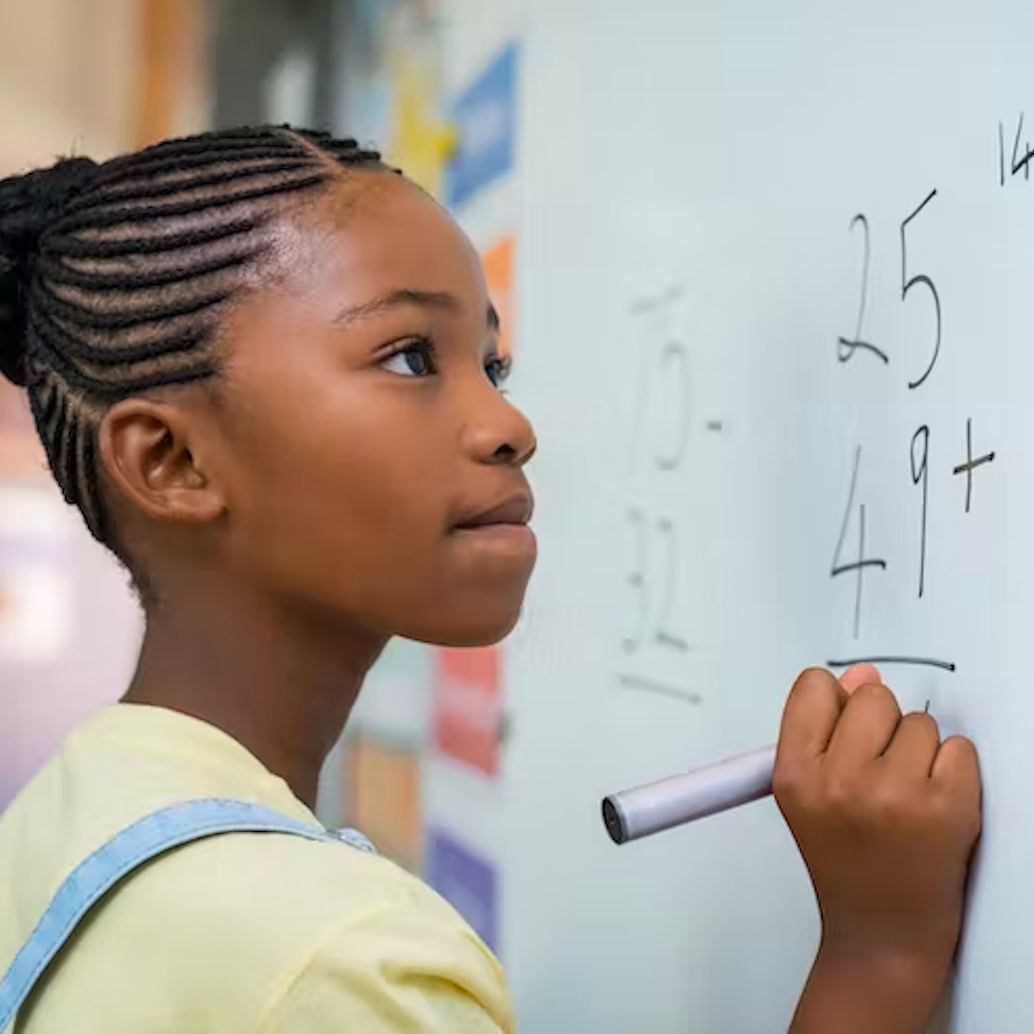}
&
\includegraphics[width=30mm,valign=m]{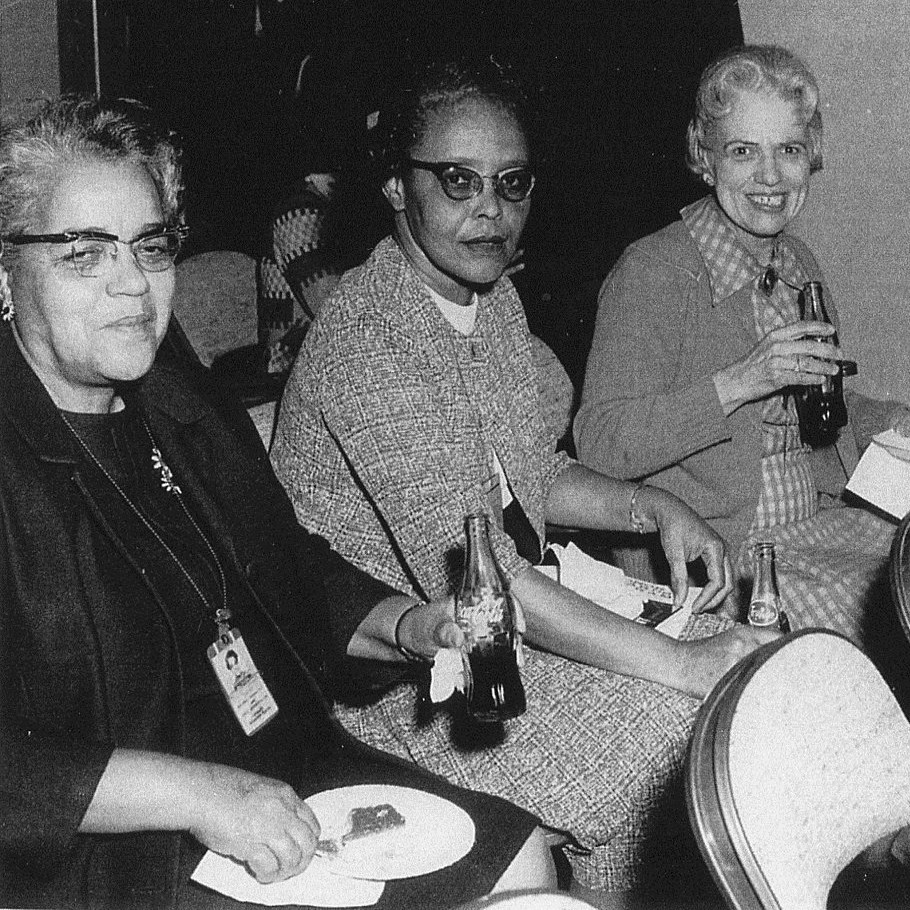}\\
%  & \includegraphics[width=30mm,valign=m, trim=0 9cm 0 6.25cm, clip]{figures/knocker-up.jpg} & % https://www.computerhistory.org/collections/catalog/102741216
%  \includegraphics[width=30mm,valign=m, trim=4cm 0 5cm 0, clip]{figures/102741216.03.01.jpg}
%  &
 % \includegraphics[width=30mm,valign=m, trim=7.5mm 0 0 0, clip]{figures/Sweatshop_in_Ludlow_Street_Tenement,_New_York_cph.3a24271.png}\\
 % https://www.flaticon.com/free-icon/sewing-machine_2175777?term=sewing&page=1&position=3&origin=style&related_id=2175777
 \textit{a}) \textit{label} & enhancement & 
 % enhancement \&
 replacement & displacement \\
 \textit{b}) \textit{valence} & beneficial & %beneficial \& 
 neutral & harmful \\
 \textit{c}) \textit{effect on cognition} & reskilling & unaffected & deskilling \\
 \textit{d}) \textit{labour obfuscation} & none & moderate & high \\
 \textit{e}) \textit{human equivalence} & same & better & quantity, not quality \\
 \textit{f}) \textit{human-in-the-loop} & abacist & none & programmer \\
 % b)& \textit{deskilling} & maybe & no & yes \\
 \textit{g}) \textit{human input} & full-blown cognition & function, numbers & code, input to code \\
 \textit{h}) \textit{desired output} & result of calculation  & result of calculation& result of calculation \\ 
 % \textit{i}) \textit{human outperformance} & no & depends & no \\
 % $\vdots$ & $\vdots$ & $\vdots$ & $\vdots$ \\
\end{tblr}

%% file: table3.tex
\definecolor{paledogwood}{HTML}{D8BDBE}
\definecolor{lavendar}{HTML}{BD90E3}
\definecolor{indianred}{HTML}{cc575f}
\definecolor{mintgreen}{HTML}{BEE3DB}
\definecolor{cambridgeblue}{HTML}{89B0AE}

\definecolor{mimipink}{HTML}{ffd6ff}
\definecolor{periwinkle}{HTML}{bbd0ff}
\definecolor{eggplant}{HTML}{533B4D}
\definecolor{champagne}{HTML}{FAE3C6}
\definecolor{hotpink}{HTML}{F564A9}
\definecolor{murray}{HTML}{990B4F}
\definecolor{mossgreen}{HTML}{87986A}
\definecolor{verdigris}{HTML}{18A3A5}

    \centering
    \sffamily%\footnotesize
    \begin{tblr}{
        colspec={X[l]X[l]X[l]X[l]X[l]},
        cells={valign=m},
        row{1}={white, halign=c},
        cell{1}{1} = {font=\itshape\bfseries},
        column{1} = {halign=l, leftsep=0pt},
        column{2-Z}={31mm},
        % column{2}={30mm},
        % column{3}={31mm},
        % column{4}={32mm},
        cell{1}{2-Z} = {font=\sc\sffamily, fg=verdigris!50!black},
        row{4-Z}= {2\baselineskip},
        cell{odd}{2} = {mintgreen!15!white},
        cell{odd}{3} = {mintgreen!50!cambridgeblue!15!white},
        cell{odd}{4} = {mintgreen!15!white},
        % row{2-3}={bg=white, halign=c},
        row{2-3}={halign=c},
        row{2}={2.5cm},
        row{3}={3.5cm},
        % row{Z}={halign=c},
        cell{1}{2}={bg=mintgreen, halign=c},
        cell{1}{3}={bg=mintgreen!50!cambridgeblue, halign=c},
        cell{1}{4}={bg=mintgreen, halign=c},
    }
    \makecell[l]{artefact \\ versus \\ cognitive labour} &
        \makecell[c]{Alarm Clock \\ versus \\ Knocker-Upper} &
            \makecell[c]{Camera \\ versus \\ Human Vision} &
                \makecell[c]{Garment Factory \\ versus \\ Seamstress/Tailor} \\
    & \includegraphics[height=18mm, valign=b]{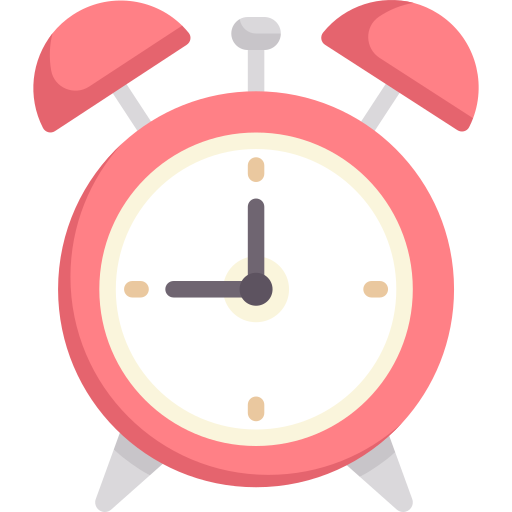} & 
        \includegraphics[height=18mm, valign=b]{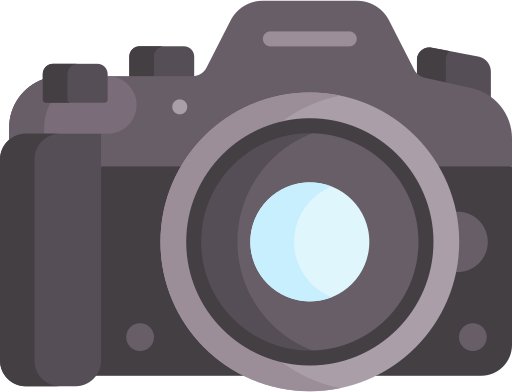} &
            \includegraphics[height=18mm, valign=b]{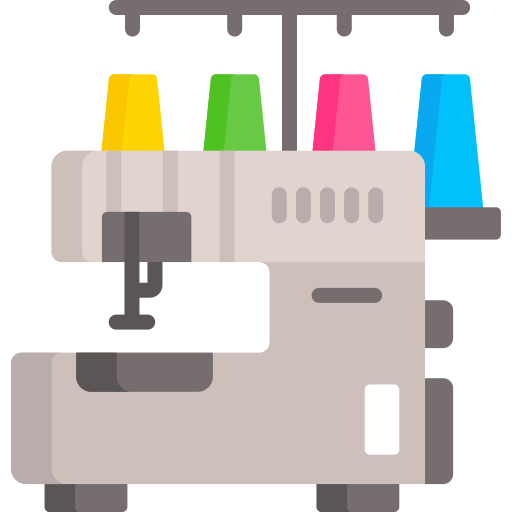} \\
    & \includegraphics[width=30mm,valign=m]{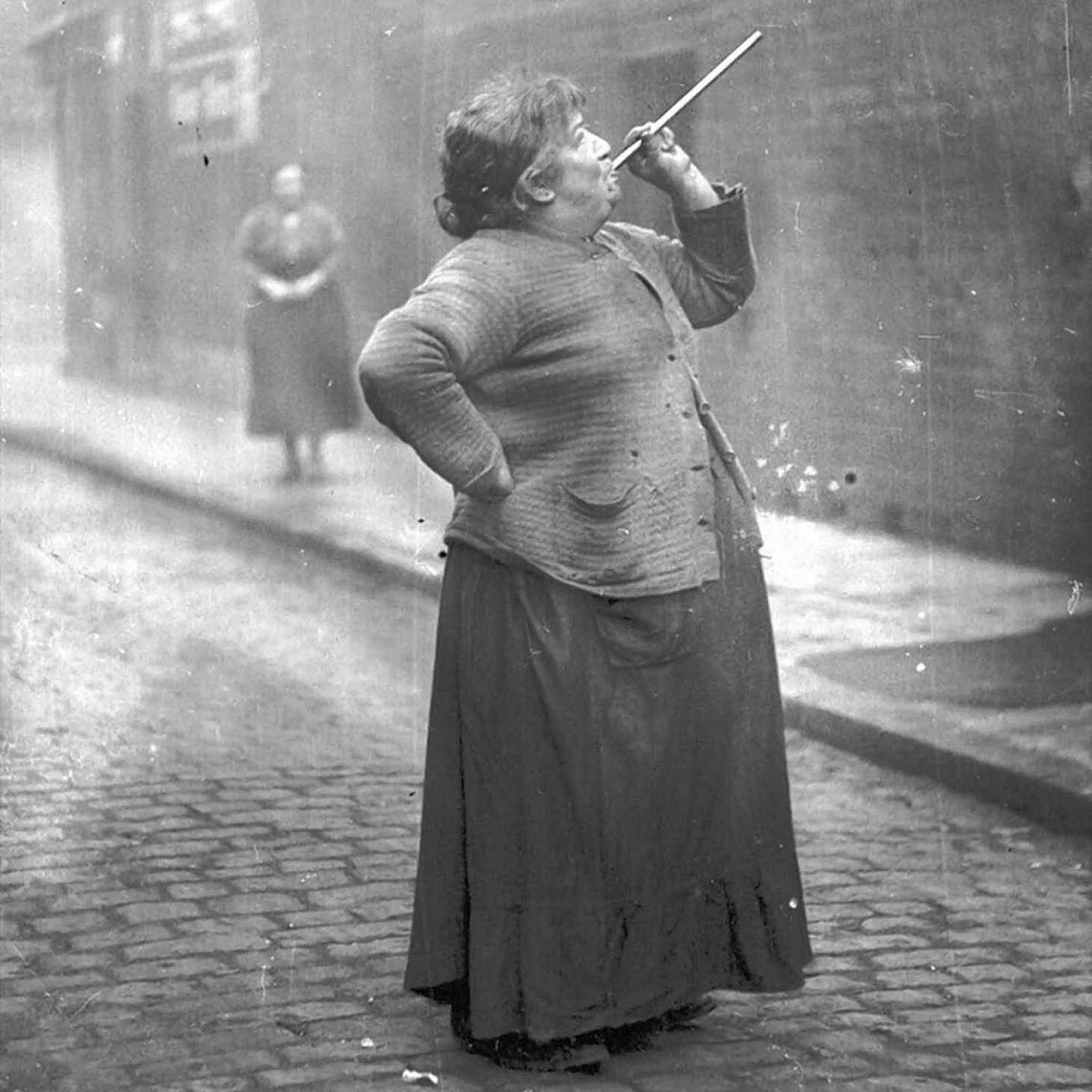} & % https://www.computerhistory.org/collections/catalog/102741216
    \includegraphics[width=30mm,valign=m]{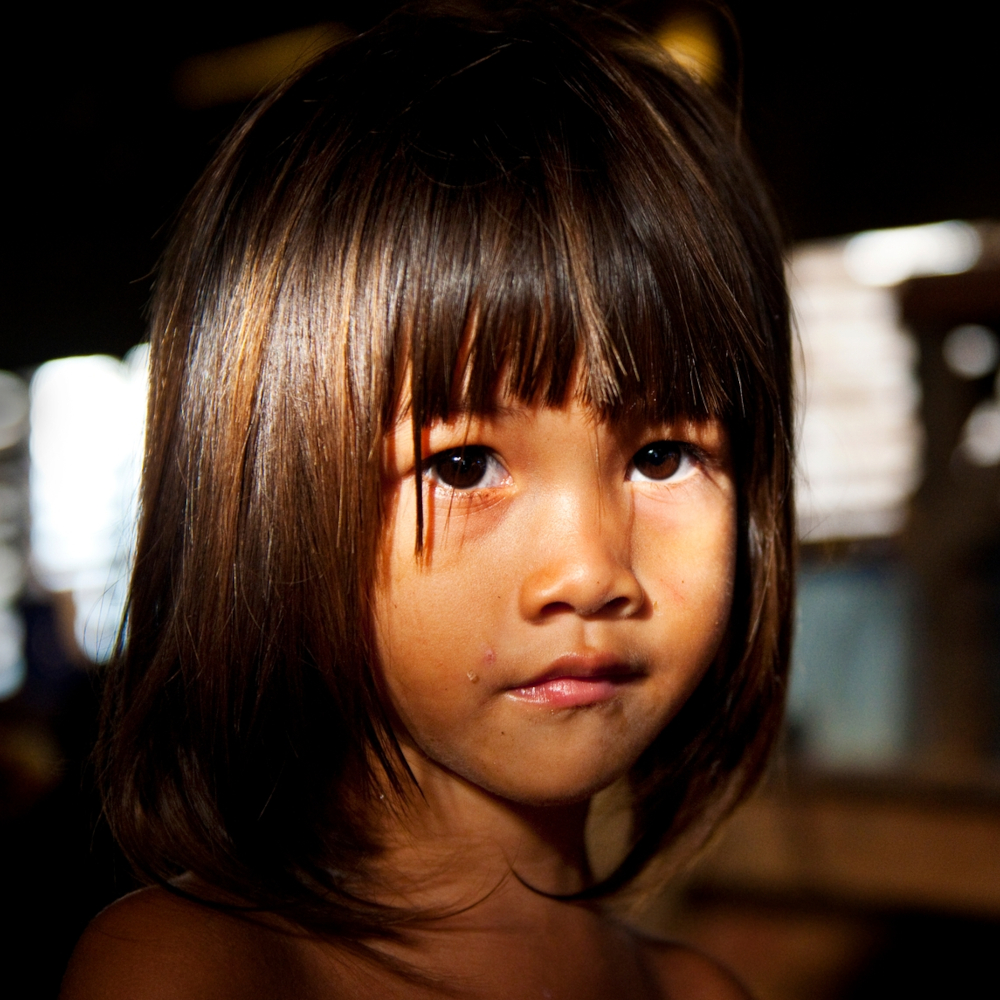}
    &
    \includegraphics[width=30mm,valign=m]{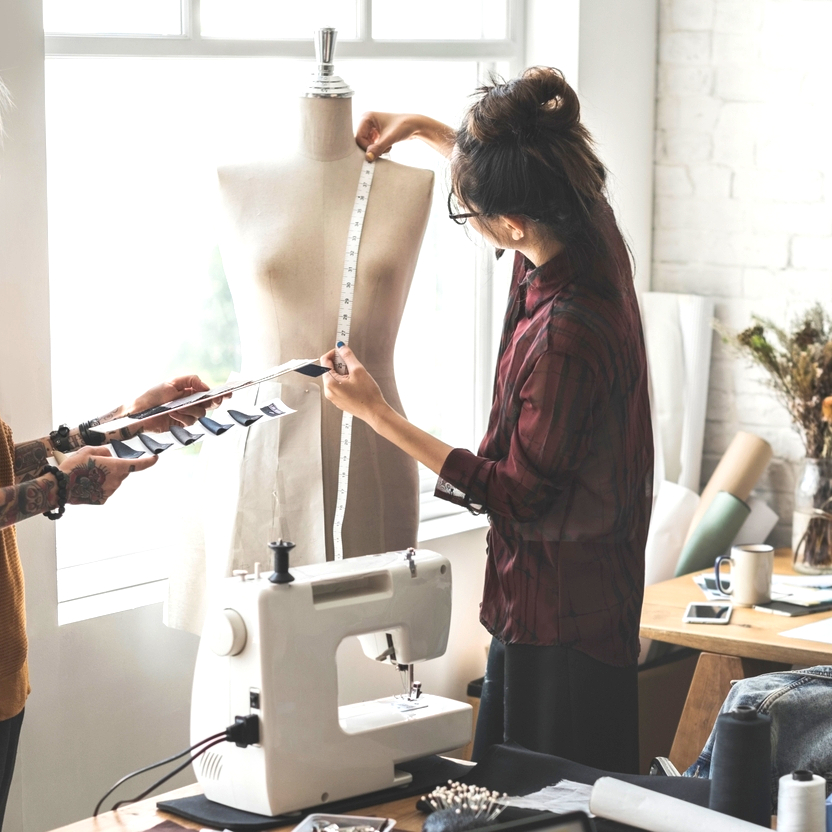}\\
      % https://www.flaticon.com/free-icon/sewing-machine_2175777?term=sewing&page=1&position=3&origin=style&related_id=2175777
    \textit{a}) \textit{label} & replacement & enhancement & displacement \\
    \textit{b}) \textit{valence} & neutral & beneficial & harmful \\
    \textit{c}) \textit{effect on cognition} & reskilling & unaffected & deskilling \\
    \textit{d}) \textit{labour obfuscation} & minimal & moderate & maximal \\
    \textit{e}) \textit{human equivalence} & worse or same & different or better & quantity, not quality \\
    \textit{f}) \textit{human-in-the-loop} & none & user & workers \\
    % b)& \textit{deskilling} & maybe & no & yes \\
    \textit{g}) \textit{human input} & current time, ring time, energy source & framing, memory or film, energy source & pattern, materials, workers, wages \\
    \textit{h}) \textit{desired output} & user awake & persistent image or video & clothes \\ 
    % \textit{i}) \textit{human outperformance} & no & depends & no \\
    % $\vdots$ & $\vdots$ & $\vdots$ & $\vdots$ \\
\end{tblr}

%% file: table4.tex
\definecolor{paledogwood}{HTML}{D8BDBE}
\definecolor{lavendar}{HTML}{BD90E3}
\definecolor{indianred}{HTML}{cc575f}
\definecolor{mintgreen}{HTML}{BEE3DB}
\definecolor{cambridgeblue}{HTML}{89B0AE}

\definecolor{mimipink}{HTML}{ffd6ff}
\definecolor{periwinkle}{HTML}{bbd0ff}
\definecolor{eggplant}{HTML}{533B4D}
\definecolor{champagne}{HTML}{FAE3C6}
\definecolor{hotpink}{HTML}{F564A9}
\definecolor{murray}{HTML}{990B4F}
\definecolor{mossgreen}{HTML}{87986A}
\definecolor{verdigris}{HTML}{18A3A5}

    \centering
    \sffamily
    \begin{tblr}{
        colspec={X[l]X[l]X[l]X[l]X[l]},
        cells={valign=m},
        row{1}={white, halign=c},
        cell{1}{1} = {font=\itshape\bfseries},
        column{1} = {halign=l, leftsep=0pt},
        column{2-Z}={31mm},
        % column{2}={30mm},
        % column{3}={31mm},
        % column{4}={32mm},
        cell{1}{2-Z} = {font=\sc\sffamily, fg=verdigris!50!black},
        row{4-Z}= {2\baselineskip},
        cell{odd}{2} = {mintgreen!15!white},
        cell{odd}{3} = {mintgreen!50!cambridgeblue!15!white},
        cell{odd}{4} = {mintgreen!15!white},
        % row{2-3}={bg=white, halign=c},
        row{2-3}={halign=c},
        row{2}={2.5cm},
        row{3}={3.5cm},
        % row{Z}={halign=c},
        cell{1}{2}={bg=mintgreen, halign=c},
        cell{1}{3}={bg=mintgreen!50!cambridgeblue, halign=c},
        cell{1}{4}={bg=mintgreen, halign=c},
    }
    \makecell[l]{artefact \\ versus \\ cognitive labour} &
        \makecell{LLM \\ versus \\ Essay Writing} &
            \makecell{Image Generator \\ versus \\ Artist} &
                \makecell{Chatbot \\ versus \\ Companionship} \\
    & \includegraphics[height=18mm, valign=m]{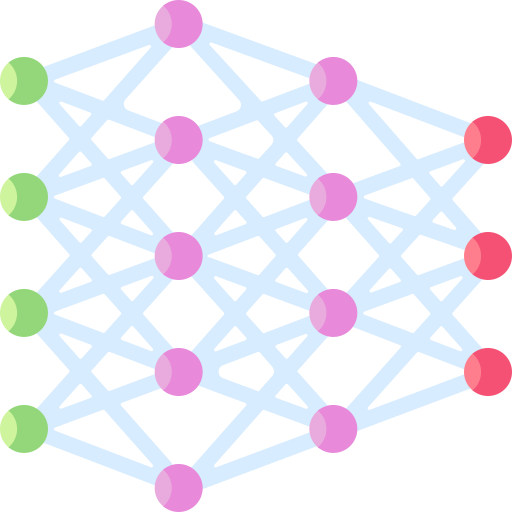} & 
            % https://www.flaticon.com/free-icon/deep-learning_9716180?related_id=9716180&origin=pack
        \includegraphics[height=18mm, valign=m]{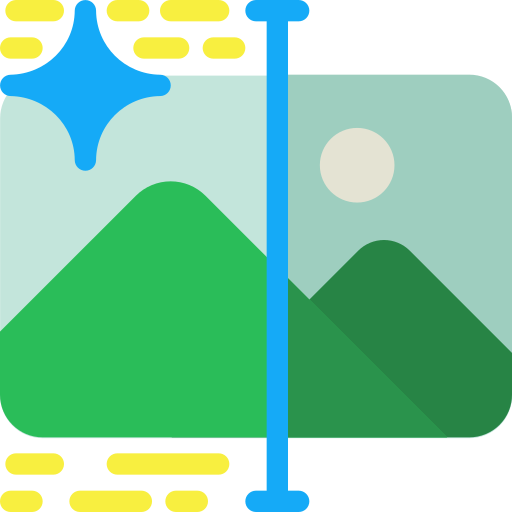} &
        % https://www.flaticon.com/free-icon/generative-image_16649259?related_id=16649259&origin=pack
            \includegraphics[height=18mm, valign=m]{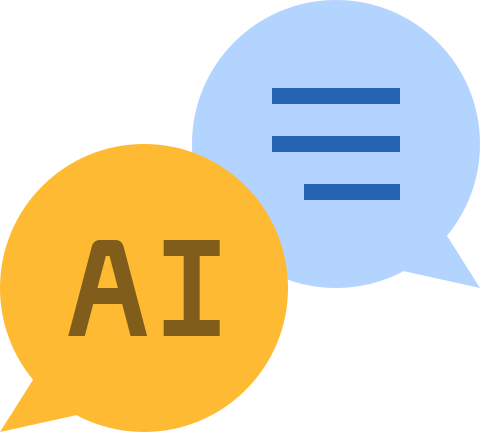} \\
%        % https://www.flaticon.com/free-icon/chatbot_12122368?term=chatbot&related_id=12122368

    & \includegraphics[width=30mm,valign=m]{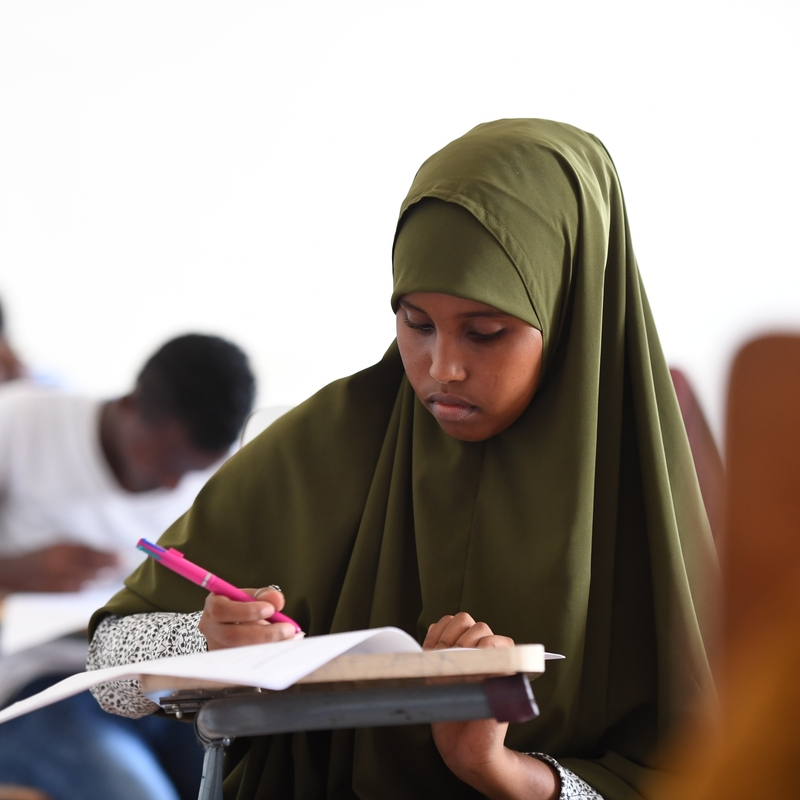} & % https://www.computerhistory.org/collections/catalog/102741216
    \includegraphics[width=30mm,valign=m]{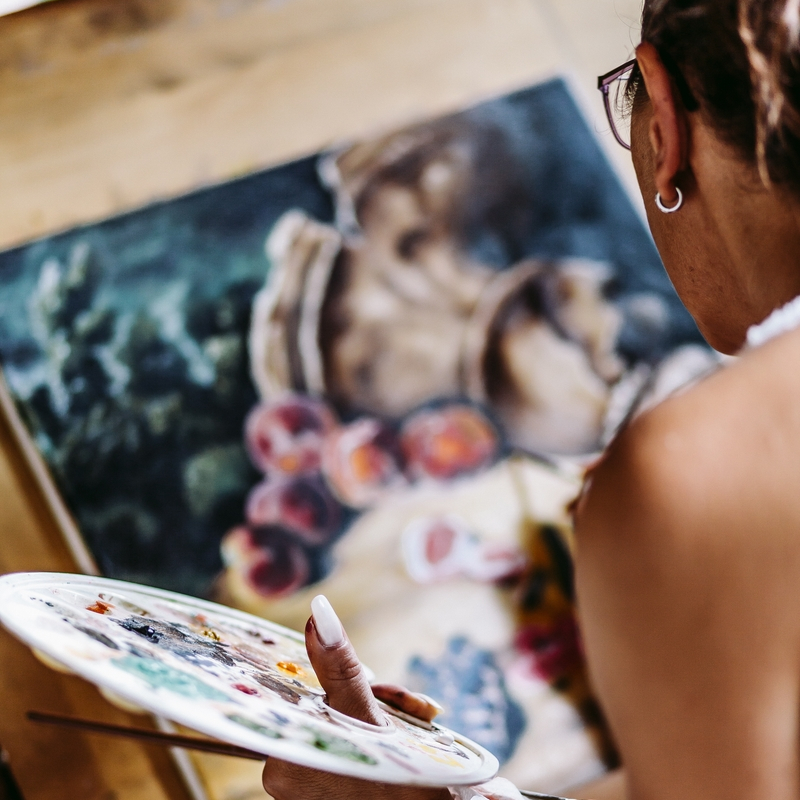}
    &
    \includegraphics[width=30mm,valign=m]{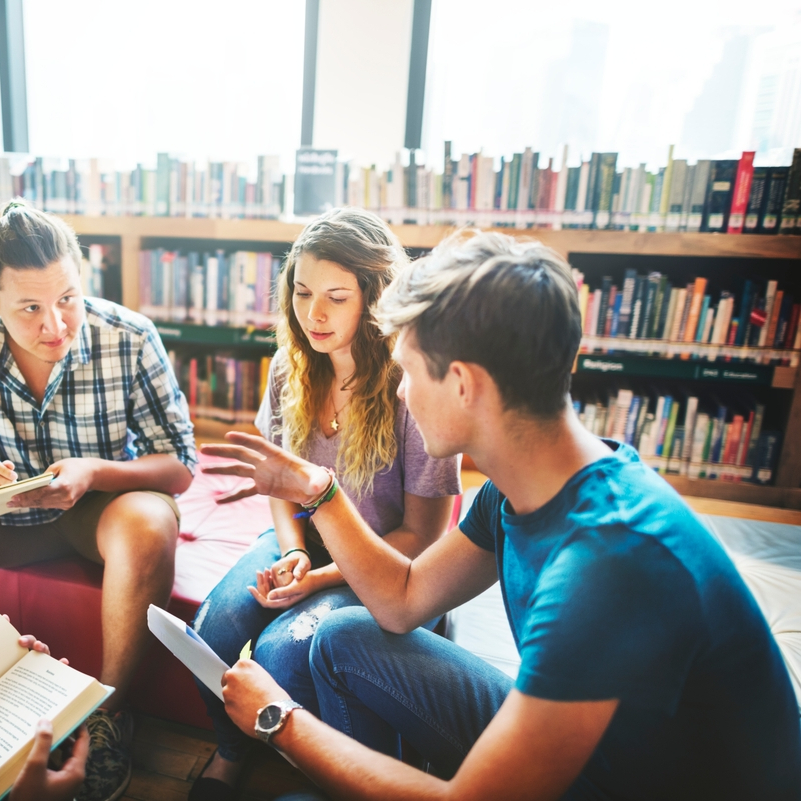}\\
      % https://www.flaticon.com/free-icon/sewing-machine_2175777?term=sewing&page=1&position=3&origin=style&related_id=2175777
    \textit{a}) \textit{label} & displacement & displacement & displacement \\
    \textit{b}) \textit{valence} & harmful & harmful & harmful \\
    \textit{c}) \textit{effect on cognition} & deskilling & deskilling & deskilling \\
    \textit{d}) \textit{labour obfuscation} & maximal & maximal & maximal \\
    \textit{e}) \textit{human equivalence} & worse & worse & worse \\
    \textit{f}) \textit{human-in-the-loop} & data, programmers, sweatshop workers & data, programmers, sweatshop workers & data, programmers, sweatshop workers\\
    % b)& \textit{deskilling} & maybe & no & yes \\
    \textit{g}) \textit{input} & so-called prompts & so-called prompts  & so-called prompts  \\
    \textit{h}) \textit{desired output} & essay matching prompt & image matching prompt  & unclear, wellness \\ 
    % \textit{i}) \textit{human outperformance} & no & depends & no \\
    % $\vdots$ & $\vdots$ & $\vdots$ & $\vdots$ \\
\end{tblr}
    